\documentclass{article} 
\usepackage{iclr2025_conference,times}

\usepackage{amsmath,amsfonts,bm}

\def\eqref#1{equation~\ref{#1}}

\def\floor#1{\lfloor #1 \rfloor}
\def\1{\bm{1}}

\DeclareMathAlphabet{\mathsfit}{\encodingdefault}{\sfdefault}{m}{sl}
\SetMathAlphabet{\mathsfit}{bold}{\encodingdefault}{\sfdefault}{bx}{n}

\usepackage{hyperref}
\usepackage{url}

\usepackage{enumitem}
\usepackage{graphicx}
\usepackage{array}
\usepackage{booktabs}
\usepackage{subcaption}
\usepackage{algorithm}
\usepackage{algpseudocode}
\usepackage{adjustbox}
\usepackage{booktabs}
\usepackage{bbm}
\usepackage{multirow}
\usepackage{mathtools}
\usepackage{amsmath}
\usepackage{caption} 
\usepackage{float}

\DeclareMathOperator{\Contribution}{\mathbf{Contribution}}
\DeclareMathOperator{\Coverage}{\mathbf{Coverage}}
\DeclareMathOperator{\ThresholdFilter}{\mathbf{ThresholdFilter}}
\DeclareMathOperator{\CacheHitRate}{\mathbf{CacheHitRate}}
\DeclareMathOperator{\TopK}{\mathbf{TopK}}
\DeclareMathOperator{\Sparsity}{\mathbf{\gamma}}

\title{VL-Cache: Sparsity and Modality-Aware KV Cache Compression for Vision-Language Model Inference Acceleration}

\author{Dezhan Tu 
\thanks{Work done during internship at Amazon AWS AI}\\
Department of Electrical and Computer Engineering\\
University of California, Los Angeles\\
Los Angeles, CA, USA \\
\texttt{dztu@g.ucla.edu} \\
\And
Danylo Vashchilenko \\
AWS AI \\
Amazon \\
New York, NY, USA \\
\texttt{vdanylo@amazon.com} \\
\And
Yuzhe Lu \\
AWS AI \\
Amazon \\
Santa Clara, CA, USA \\
\texttt{yuzhelu@amazon.com} \\
\And
Panpan Xu \\
AWS AI \\
Amazon \\
Santa Clara, CA, USA \\
\texttt{xupanpan@amazon.com}\\
}

\newcommand{\vlcache}{VL-Cache}

\iclrfinalcopy 

\begin{document}

\maketitle

\begin{abstract}

Vision-Language Models (VLMs) have demonstrated impressive performance across a versatile set of tasks. A key challenge in accelerating VLMs is storing and accessing the large Key-Value (KV) cache that encodes long visual contexts, such as images or videos. While existing KV cache compression methods are effective for Large Language Models (LLMs), directly migrating them to VLMs yields suboptimal accuracy and speedup. To bridge the gap, we propose \vlcache, a novel KV cache compression recipe tailored for accelerating VLM inference. In this paper, we first investigate the unique sparsity pattern of VLM attention by distinguishing visual and text tokens in prefill and decoding phases. Based on these observations, we introduce a \textit{layer-adaptive sparsity-aware cache budget allocation} method that effectively distributes the limited cache budget across different layers, further reducing KV cache size without compromising accuracy. Additionally, we develop a \textit{modality-aware token scoring policy} to better evaluate the token importance. Empirical results on multiple benchmark datasets demonstrate that retaining only 10\% of KV cache achieves accuracy comparable to that with full cache. In a speed benchmark, our method accelerates end-to-end latency of generating 100 tokens by up to 2.33x and speeds up decoding by up to 7.08x, while reducing the memory footprint of KV cache in GPU by 90\%.

\end{abstract}

\newcommand{\panpan}[1]{{\color{orange}{PX: #1}}}
\newcommand{\huan}[1]{{\color{green}{Huan: #1}}}
\newcommand{\bryan}[1]{{\color{red}{BL: #1}}}
\newcommand{\dezhan}[1]{{\color{blue}{dz: #1}}}
\newcommand{\vdanylo}[1]{{\color{purple}{D: #1}}}
\newcommand{\handing}[1]{{\color{teal}{Han: #1}}}
\newcommand{\todo}[1]

\section{Introduction}
\label{sec:introduction}

Vision-Language Models (VLMs) have recently emerged as powerful tools for a broad range of multi-modal tasks \citep{liu2023llava, chen2023pali, bai2023qwen}. As these models improve at processing long visual context -- such as high-resolution images, multiple images and multi-frame videos \citep{li2024llava-onevision} -- the number of visual tokens increase rapidly. Consequently, deploying VLMs demands substantial GPU memory capacity, bandwidth, and computational resources, leading to high inference latency and cost.

Similarly to Large Language Models (LLMs) \citep{chang2024survey}, VLMs decode tokens sequentially in an auto-regressive loop. The key and value pairs of the input prompt and of the generated output tokens are stored in GPU memory (where they are known as the KV cache) and reused at each decoding step to avoid recomputation. As the context length grows, KV cache not only occupies a larger amount of GPU memory, but also increases inference latency due to data movement between GPU's high-bandwidth memory (HBM) and its on-chip memory (SRAM) in each decoding step \citep{dao2023flashattention, hong2024flashdecoding++}. This is a significant challenge for scaling VLMs, because large KV cache is required to hold the input images and video frames. For example, with a batch size of four prompts, five images in each prompt, and each image using 2K visual tokens, serving the LLaVA-1.6-34B model requires 110 GB of HBM capacity just for the KV cache of visual tokens.  

\begin{figure}[htbp]
    \centering
    \hfill
    \begin{minipage}[b]{\textwidth}
        \centering
        \begin{subfigure}[b]{0.49\textwidth}
            \centering
            \includegraphics[width=\textwidth]{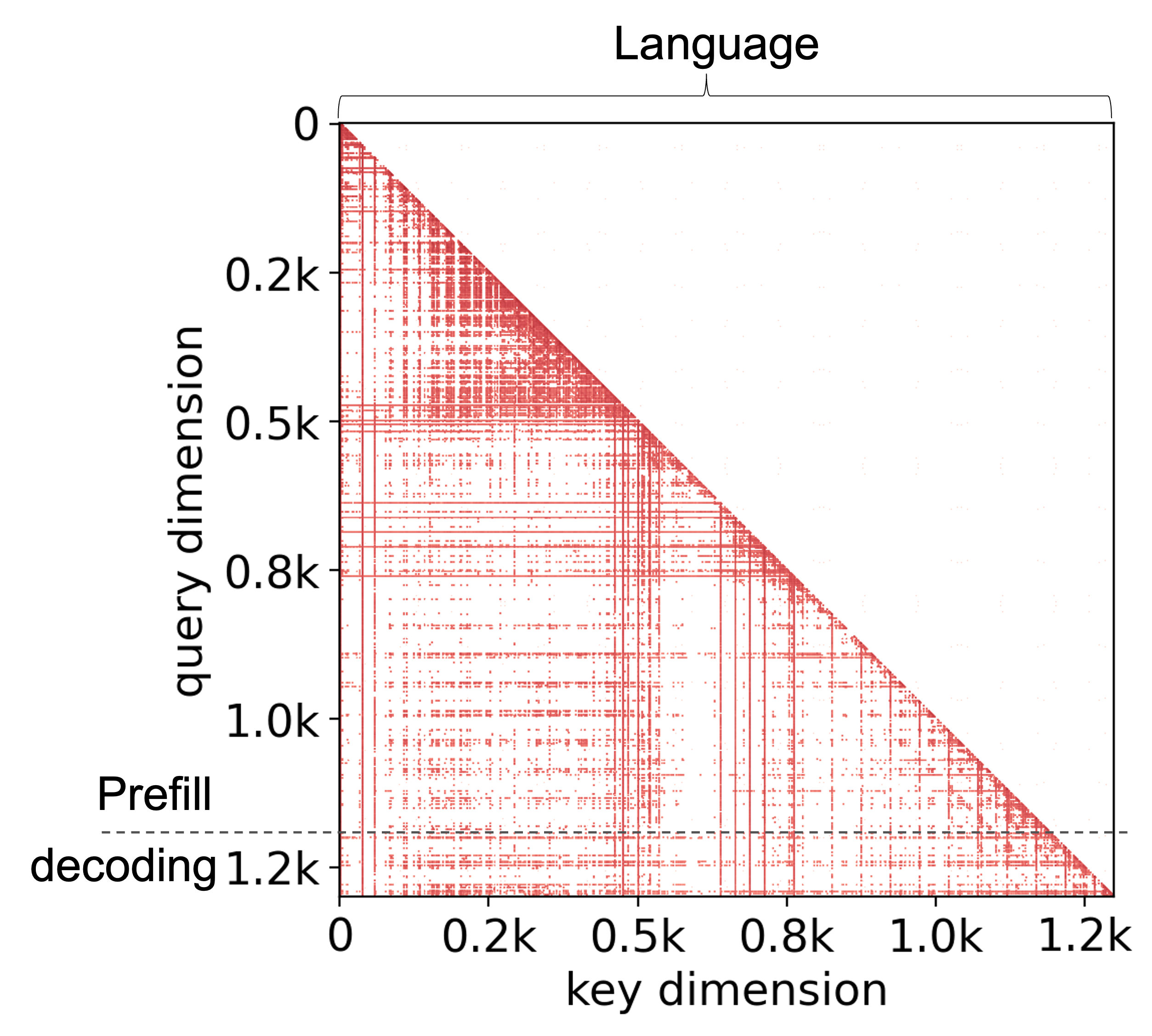}
            \caption{LLMs Attention Score Matrix}
            \label{fig:llm_attn}
        \end{subfigure}
        \hfill
        \begin{subfigure}[b]{0.48\textwidth}
            \centering
            \includegraphics[width=\textwidth]{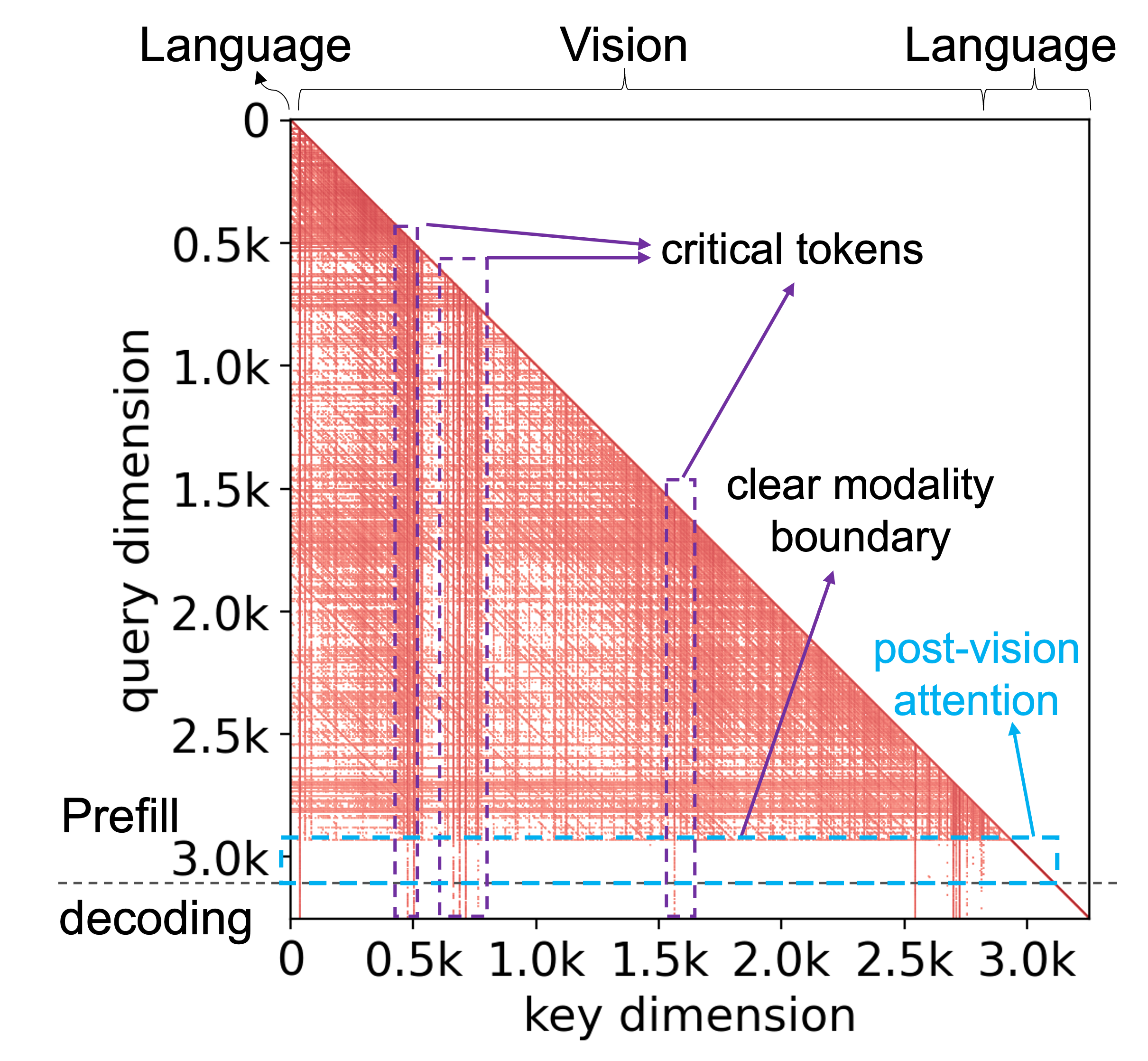}
            \caption{VLMs Attention Score Matrix}
            \label{fig:vlm_attn}
        \end{subfigure}
        \caption{Attention Score Matrix from LLaVA v1.6 Mistral 7B with (a) language-only context, and (b) language and vision context. A deeper red color indicates a higher attention score. Both matrices indicate that critical tokens used in the decoding phase are primarily consistent with those in the prefill context. The key difference is that, in (b) VLMs attention, a clear modality boundary emerges along the query dimension.}
        \label{fig:llm_vlm_attn}
    \end{minipage}
\end{figure}

KV cache sparsification is a promising method of reducing GPU memory requirements and inference latency for LLMs \citep{zhang2024h2o,zhang2024pyramidkv,yang2024pyramidinfer,he2024zipcache,li2024snapkv}. Prior methods leveraged sparsity in the attention scores matrix of LLMs to evict insignificant tokens from KV cache, while preserving important contextual information needed for output token generation. Besides reducing the GPU memory footprint, this method also reduces the token generation latency by minimizing the amount of data movement and floating-point operations (FLOPs) in the attention layer. 

We discover that VLMs and LLMs exhibit significantly different attention sparsity patterns, as illustrated in \autoref{fig:llm_vlm_attn}. In VLMs, a clear modality boundary exists between visual tokens and the subsequent language tokens. Specifically, the attention patterns from the output language tokens is much closer aligned with the language tokens that follow the visual tokens in the prompt (the \textbf{post-vision attention}) rather than the visual tokens themselves. Previous KV compression methods are modality-unaware and incorrectly conflate these two sources of attention scores. Unsurprisingly, our experiment in Figure \ref{fig:accuracy} shows that KV cache compression methods designed for LLMs have suboptimal results when applied to VLMs. 

Next, we measure that attention sparsity ratios vary in 70\% to 99\% range across transformer layers. Our observations in Section \ref{observation:layer-wise sparsity} show that previous KV compression methods suboptimally distribute the cache budget between layers, which leads to either under- or over-compression of information in KV cache. Finally, we observe that sparsity ratios differ between visual and language tokens as well, so an optimal cache budget allocation between layers can not be done before sparsity is measured in a particular prompt.

To address these gaps, we propose \vlcache, a novel KV cache compression method for accelerating \textbf{V}ision-\textbf{L}anguage Model inference. Our method fully utilizes cross-modality and layer-wise attention sparsity patterns in VLMs to dynamically prune KV cache with minimal loss of task-level accuracy. To the best of our knowledge, this is the first work that investigates attention sparsity in VLMs and the first work that specifically optimizes KV cache compression for VLMs. In particular, we propose (1) sparsity-aware KV cache budget allocation between the transformer layers at inference time, and (2) modality-aware scoring policy for token eviction. 

Experimental results in Section \ref{acc_eval} show that our method retains 98\% of the original task-level accuracy while using only 10\% of KV cache for the majority of vision-language tasks from several datasets. In a speed benchmark, our method reduces end-to-end latency of generating 100 tokens by up to \textbf{2.33x} and speeds up decoding in particular by up to \textbf{7.08x}, while allocating \textbf{90\% less} GPU memory for KV cache. In inference scenarios where KV cache size is the limiting factor to higher concurrency, \vlcache\ enables up to \textbf{10x} higher concurrency after KV cache compression. 

Overall, our contributions in this paper are summarized as follows:

\begin{itemize}[leftmargin=0.4cm]
\item \textbf{VLMs Attention Sparsity Profile.} We uncovered unique attention sparsity patterns in the prefill and decoding phases of VLM inference on a variety of multi-modal tasks, which is drastically different from those of LLMs.
\item \textbf{Layer-Adaptive Sparsity-Aware Cache Budget Allocation.} We proposed to allocate each layer's KV cache budget with consideration of that layer's attention sparsity at inference time. Even with 10\% KV cache budget, we retained high accuracy on popular benchmark tasks.
\item \textbf{Modality-aware Token Scoring Policy.}  We observed that language-to-vision attention robustly includes information about the importance of visual tokens. We treat visual and language attention scores differently to better retain important tokens.
    
\end{itemize}

\section{Background}

In this section, we detail the inference procedure of the widely adopted VLM architecture with vision \& language input and text output, where image tokens are projected as soft prompts \citep{li2024llava-onevision, team2023gemini, islam2024gpt, bai2023qwen, anthropic2023claude}. We also introduce the formulation of KV cache compression and the approach of existing works \citep{zhang2024h2o, zhang2024pyramidkv, liu2024scissorhands, ge2023model, tang2024quest, li2024snapkv, lee2024infinigen, yu2024effectively}. 

\subsection{VLM inference}

\paragraph{Prefill phase.} The input to VLMs includes both images and language, where images are processed by the visual encoder to generate visual tokens. Subsequently, a projection layer, such as a simple multi-layer perceptron, maps these visual tokens to a unified embedding space. Meanwhile the language prompt is fed to a tokenizer and embedding layer to create the initial hidden state for language tokens. For notation simplicity, we denote a sequence of $m$ prompt tokens, including both visual and language ones, as $\{x_1, ..., x_m\}$. These tokens are processed by the language model in parallel to calculate the probability of the first decoded token $P_{\theta}(x_{m+1} | x_1, ..., x_m)$. Simultaneously, the key vectors $\{k_1^{(l)}, ..., k_m^{(l)}\}$ and value vectors $\{v_1^{(l)}, ..., v_m^{(l)}\}$ at each transformer layer $l$ are cached in GPU memory to avoid recomputation in the next decoding step.

\paragraph{Decoding phase.} Once decoding starts, the language model in VLMs takes effect and generates one token per step in an auto-regressive loop. At step $i$, the model receives the token $x_{m+i}$ as input and calculates the probability $P_{\theta}(x_{m+i+1} | x_1, ..., x_{m+i})$. Each decoding step involves generating new key vectors $k_{m+i}^{(l)}$ and value vectors $v_{m+i}^{(l)}$, which are appended to the previously cached key-value pairs for each layer, resulting in $\{k_1^{(l)}, ..., k_{m+i}^{(l)}\}$ and $\{v_1^{(l)}, ..., v_{m+i}^{(l)}\}$. In case of long contexts, such as multiple or high resolution images, the key-value cache can grow significantly larger than the model parameters and other intermediate tensors, making memory capacity and bandwidth major performance bottlenecks.

\subsection{KV Cache Compression}
To address the bottleneck of storing and accessing the large KV cache during decoding, many researchers have focused on KV cache compression to maintain only a subset of the full KV cache for more efficient decoding while minimizing the accuracy loss. There are two main design dimensions to such algorithms: how many cache tokens should be kept at each layer, and which tokens to evict during compression. 
\paragraph{Budget Allocation.}
Since the transformer architecture consists of multiple identical layers, a straightforward strategy is to allocate an equal budget of KV cache slots to each layer \citep{xiaoefficient,zhang2024h2o, he2024zipcache}. More recently, inspired by the observation that removing cache tokens at different layers results in varying degrees of performance loss, PyramidKV \citep{zhang2024pyramidkv} and PyramidInfer \citep{yang2024pyramidinfer} proposed a decay schedule to assign more cache budget to shallower layers and observed improved accuracy results relative to the baseline with equal budget per layer.

\paragraph{Token Scoring Policy.}
For a given cache token budget, a token scoring policy $\psi$ is required to rank the importance of KV cache tokens and decide which tokens to keep. Let $n$ be the count of cache tokens at layer $l$, and $S = \{0, 1, ..., n\}$ be the indices of these cache tokens. We define $\psi : S \rightarrow \mathbb{R}^{n}$, with the output being the scores for current cache tokens. Given these scores, indices with top-$k$ scores, $S_{\psi} \coloneqq \{i_1, i_2, ..., i_k: \psi(S)_{i_j} \geq \psi(S)_{x \in [n]\backslash \{i_1, i_2, ..., i_k\}} \}$, where $k \in [1, n)$, are selected. Recent works have shown that attention scores serve as an effective source to design such policies. StreamingLLM \citep{xiaoefficient} found that high attention scores are frequently assigned to initial tokens and achieved length generalization by keeping only the initial and most recent cache tokens while removing intermediate ones. H2O \citep{zhang2024h2o} uses accumulated attention scores to identify crucial tokens to retain. 

We make the observation that, fundamentally, KV cache compression methods only work because inference with a transformer layer is a sparse process. Therefore, in this work, we leverage attention sparsity as the unified guidance to design both the cache budget allocation mechanism and the token eviction policy for KV cache compression in VLMs. 

\section{Preliminary Experiment}

The attention mechanism for visual and language tokens is a key aspect of VLMs. Therefore, further optimization of the KV cache compression methods for VLMs requires careful investigation of the attention patterns with relevant input prompts. Motivated by this need, we conducted preliminary experiments to explore the VLM attention. We randomly sampled data from three multi-modal datasets --- DocVQA \citep{mathew2021docvqa}, MathVista \citep{lu2023mathvista}, and Coco-Caption \citep{chen2015microsoft}. These datasets cover of a wide range of visual tasks, including OCR, visual diagram reasoning, and world knowledge. We selected one of the state-of-the-art VLMs --- LLaVA-Mistral-7B \citep{liu2023llava} --- and recorded the attention score matrix until the generation process completes. Initially, we investigate attention sparsity across layers, and then propose two metrics to quantitatively assess how the model attends to visual and language tokens in each layer and evaluate different token scoring policies. We conclude our main findings in the end of this section which lead to our algorithm design described in Section \ref{sec:vlcache_method}.

\subsection{Measuring Attention Sparsity}
\label{observation:layer-wise sparsity}
In this section, we measure the sparsity of the attention score matrix in different transformer layers during the prefill and decoding phases. First, we apply a filter with a relative threshold $p$ to the attention score matrix $A$:

\begin{equation}
\label{eq:sparsify}
    \ThresholdFilter(A, p)_{ij} = \begin{cases} 
      A_{ij} & \;\;\;\text{if} \; A_{ij} \geq p \cdot \max_{j}(A_{ij}) \\
      0 & \;\;\; \text{otherwise}
   \end{cases}
\end{equation}

where threshold $p \in (0, 1)$ controls the strength of induced sparsification, following \cite{zhang2024h2o}. We also heuristically set $p = 1\%$, such that the filtered-out scores have an insignificant impact on the output of the transformer layer. As will be discussed in Section \ref{kv_cache_budget_allocation}, we also need to measure sparsity at inference time, so we prefer a metric with faster runtime. $\ThresholdFilter$ is asymptotically faster than alternative methods of truncating a distribution (such as top-p and top-k), since it does not require the attention scores to be sorted. After filtration, we calculate sparsity $\Sparsity^{(l)} \in [0, 1]$ of layer $l$ as count of zero entries, normalized by the size of the lower triangular portion of the attention scores matrix:

\begin{equation}
\Sparsity^{(l)} \coloneq \frac{\sum_{i \geq j}\mathbbm{1}\lbrack{\ThresholdFilter(A^{(l)}, p)_{ij} = 0}\rbrack}{|\{A_{ij}^{(l)}: i \geq j\}|}
\label{eq:sparsity}
\end{equation}

\begin{figure}[h]
    \centering
    \begin{minipage}{0.68\textwidth}
        \centering
        \begin{subfigure}{0.5\textwidth}
            \centering
            \includegraphics[width=\linewidth]{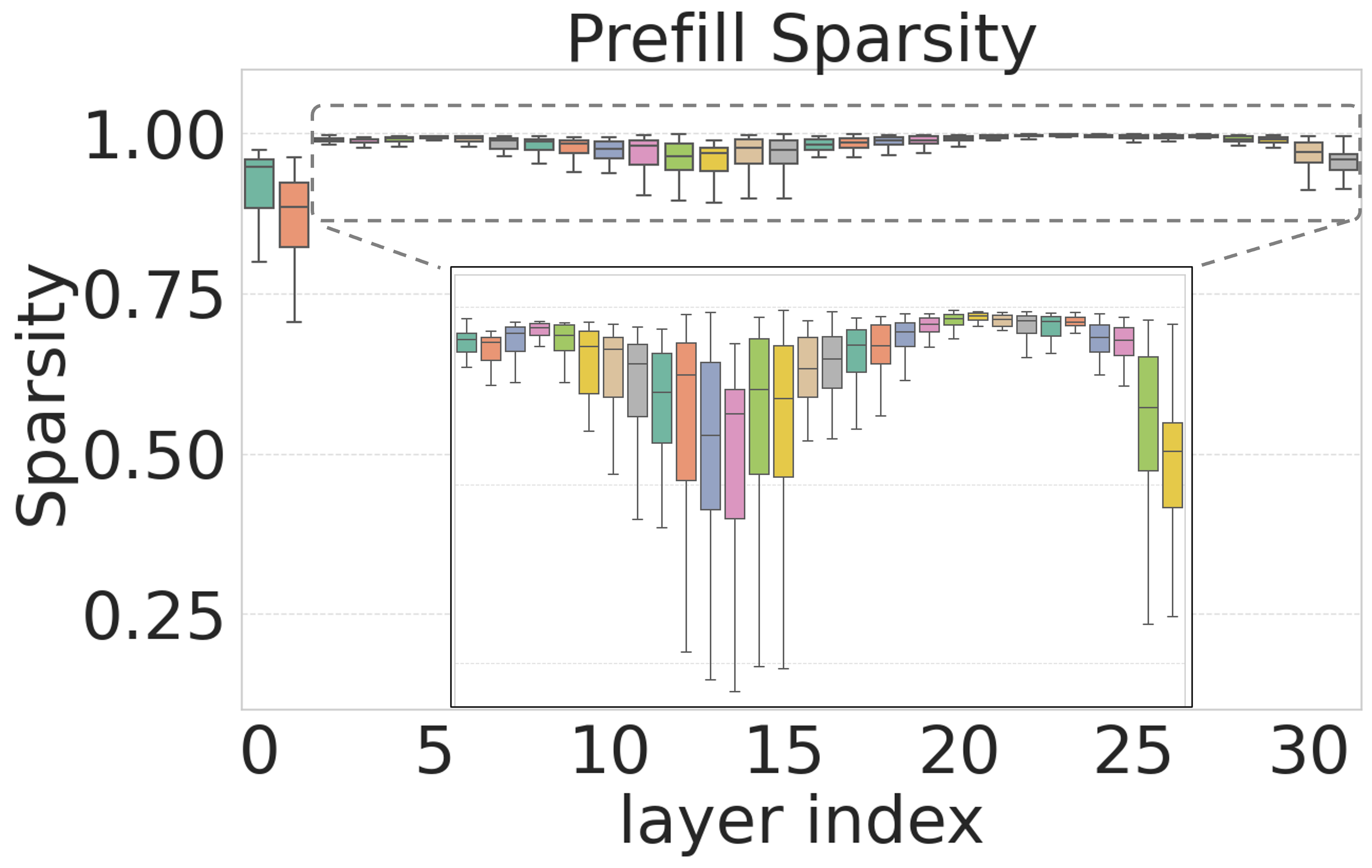}
            \caption{Prefill Sparsity}
            \label{fig:prefill_sparsity}
        \end{subfigure}%
        \hfill
        \begin{subfigure}{0.5\textwidth}
            \centering
            \includegraphics[width=\linewidth]{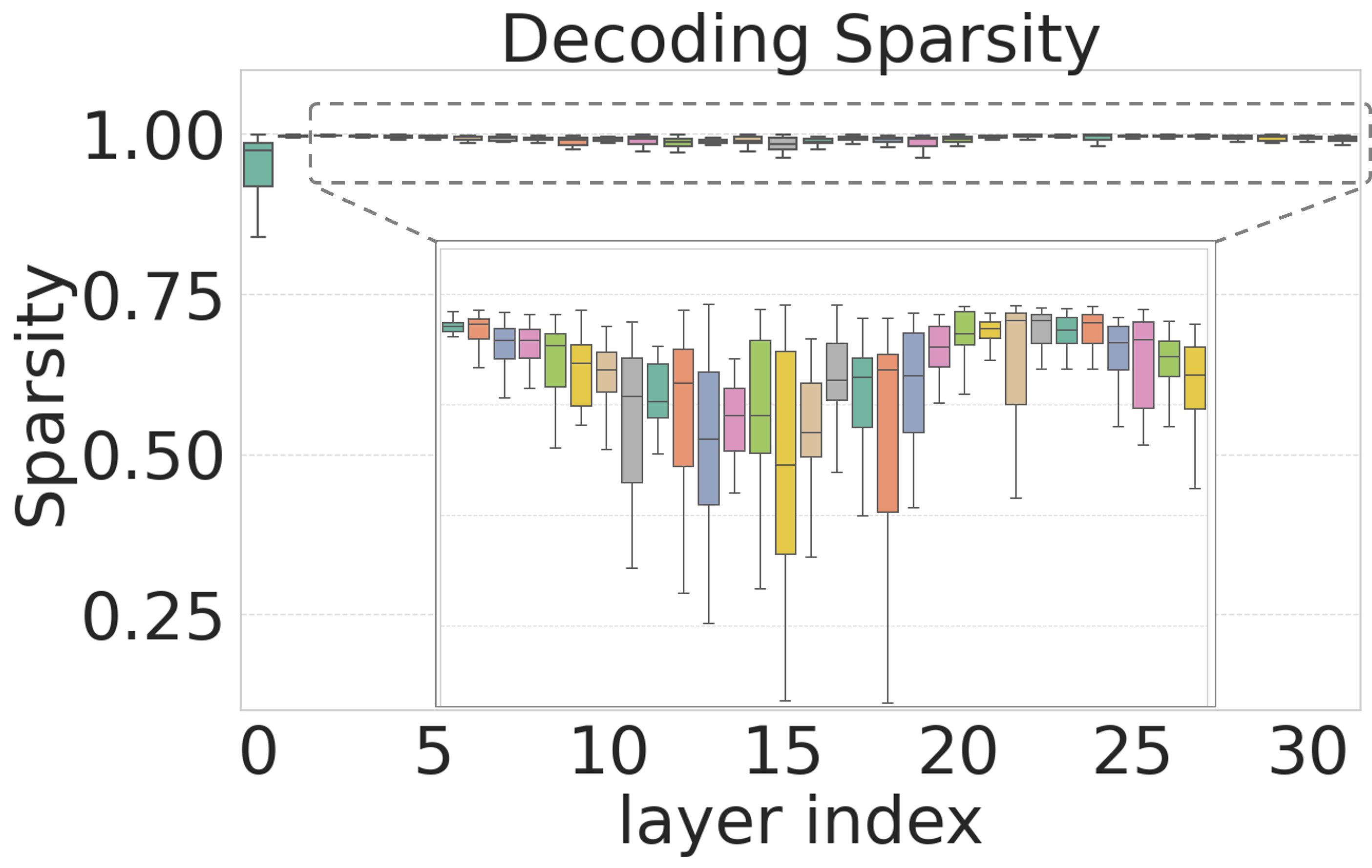}
            \caption{Decoding Sparsity}
            \label{fig:decoding_sparsity}
        \end{subfigure}
        \caption{Layer-wise attention sparsity in prefill and decoding phases. Different layers exhibit varying degrees of sparsity; The layer-wise sparsity trend in the decoding phase is similar to that in the prefill phase. }
        \label{fig:first_figure}
    \end{minipage}%
    \hfill
    \begin{minipage}{0.31\textwidth}
        \centering
        \includegraphics[width=0.98\linewidth]{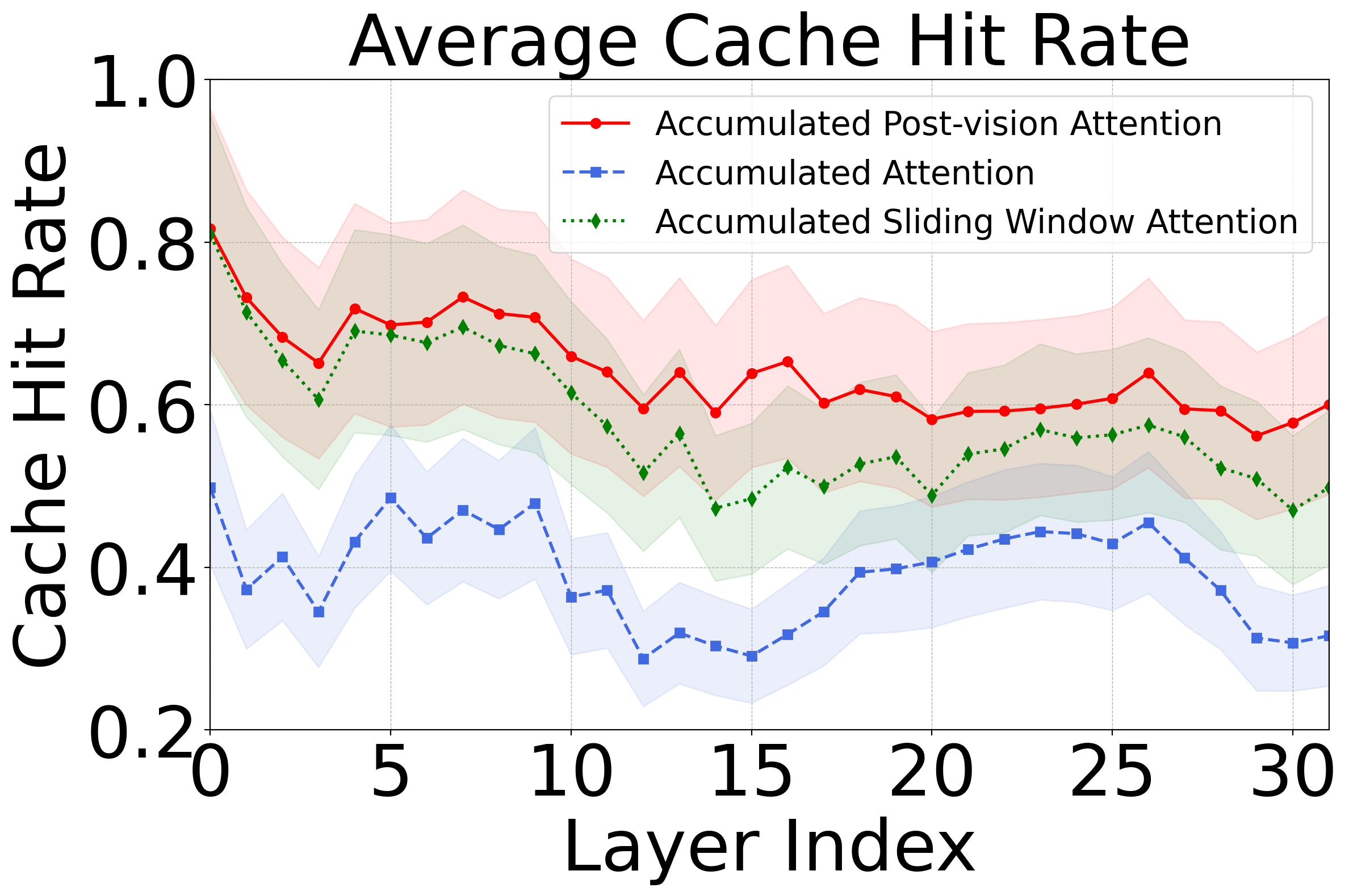}
        \caption{Average Cache Hit Rate. The accumulated post-vision attention (ours) demonstrates a higher cache hit rate compared to the other two token scoring policies.}
        \label{fig:cachehitrate}
    \end{minipage}
\end{figure}

We calculated the average sparsity over a sample from the evaluation datasets, and then plotted the range of each layer's sparsity with different attention heads in Figure \ref{fig:first_figure}. We observe that in the prefill phase (Figure \ref{fig:prefill_sparsity}), the first two layers exhibit significantly lower sparsity and higher density compared to other layers. Additionally, some layers in the middle also demonstrate higher density than their neighboring layers. During the decoding phase (Figure \ref{fig:decoding_sparsity}), a similar trend is observed as in the prefill phase, except for the second layer, which becomes much more sparse. Moreover, the Pearson Correlation Coefficient \citep{james2013introduction} is used to quantitatively compare the shape of layer-wise sparsity curves, resulting in an average shape similarity of 0.695, indicating that the curves are highly similar in both trend and magnitude. Based on these observations, we expect that the aggregate attention sparsity during the prefill phase could effectively predict the required KV cache size for robust decoding in each layer of the transformer model. 

Previous methods, such as H2O \citep{zhang2024h2o} and Keyformer \citep{adnan2024keyformer}, allocate the same cache size across all layers, leading to insufficient allocation for high-information layers and wasteful allocation for low-information layers. In a more recent work, PyramidKV \citep{zhang2024pyramidkv} and PyramidInfer \citep{yang2024pyramidinfer} monotonically decrease the KV cache size with depth of each layer, which we find to be suboptimal as well. Our observations reveal a more nuanced sparsity pattern, where different layers demand non-monotonically varying cache sizes to retain context information during decoding. We propose that sparsity based on $\ThresholdFilter$ can serve as a reliable indicator for efficient cache allocation, allowing us to leverage insights from the prefill phase at inference time to optimize KV cache allocation for the decoding phase.

\subsection{Measuring Cache Hit Rate}
\label{sec: cachehitrate}
In Figure \ref{fig:vlm_attn} of Section \ref{sec:introduction}, we previously noted a distinct boundary in VLM attention between vision and language tokens in the prompt. Based on these observations, we hypothesize that retaining tokens based on post-vision prefill attention instead of the full-prompt prefill attention would preserve important cache tokens with higher recall. To validate this hypothesis, we define $\CacheHitRate$ to measure the fraction of important tokens that are preserved after eviction when different scoring policies are applied. Let $m$ be the number of prompt tokens, $Q_{:m}, K_{:m} \in \mathbb{R}^{m \times d}$ be the corresponding query and key matrices, and $Q_{m+1} \in \mathbb{R}^{d}$ be the query vector of the first decoding token.

\paragraph{Definition 3.1}(CacheHitRate), given $A_{m+1} \coloneq softmax(\frac{Q_{m+1}{K_{:m}}^{T}}{\sqrt{d}}) \in \mathbb{R}^{m}$, we define: 

\begin{itemize}[leftmargin=*]
  \item $\psi^{*}: S \rightarrow A_{m+1}$ as the optimal scoring function based on the true attention scores in decoding;
  \item $S_{\psi^{*}}$ as the top-$k$ tokens in the order given by $\psi^{*}$, which makes it the optimal policy for any $k$;
  \item $\CacheHitRate$ as $\frac{|S_{\psi} \cap S_{\psi^{*}} |}{|S_{\psi^{*}}|}$, the percentage of $S_{\psi^{*}}$ that is also preserved by $S_{\psi}$, which denotes indices of cache tokens kept by any policy $\psi$.
\end{itemize}

We note that during inference, $A_{m+1}$ is not available because we aim to compress KV cache before the decoding pass to lift the memory bottleneck and boost decoding throughput. Thus, multiple $\psi$s have been proposed to approximate $\psi^{*}$ in recent works.

\textit{Accumulated Attention} (prior work) uses the cumulative attention score along the query dimension, $\psi(S) \coloneq \sum_{i}softmax(\frac{Q_{:m}{K_{:m}}^{T}}{\sqrt{d}})_{i}$ to rank the importance of cache tokens.

\textit{Accumulated Sliding Window Attention} (prior work) also computes accumulated attention scores along the query dimension but only over a recent window. This policy is defined as $\psi(S) \coloneq \sum_{i}softmax(\frac{Q_{m-w:m}{K_{m-w:m}}^{T}}{\sqrt{d}})_{i}$, where $w$ is a fixed window size. 

\textbf{Accumulated Post-vision Attention} (ours) is based on the observation of the similarity between attention from post-vision prompt tokens and from decoding tokens as shown in Figure \ref{fig:vlm_attn}. We adopt the term post-vision to distinguish language tokens that follow visual tokens from instructions that may precede the images in the prompt. Formally, we use the portion of the attention scores $Q_{m-\tau:m}$, where $\tau$ is the count of language tokens that follow the vision tokens in the prompt. Our policy is defined by $\psi(S) \coloneq \sum_{i}softmax(\frac{Q_{m-\tau:m}{K_{m-\tau:m}}^{T}}{\sqrt{d}})_{i}$. During prefill, one can re-interpret our policy as a dynamic prompt-dependent sliding window with window size set to the length of the post-vision language prompt instead of a static prompt-independent value. From Figure \ref{fig:cachehitrate}, we find our scoring policy leads to consistently higher cache hit rate across layers than other policies.

\section{VL-Cache Method}
\label{sec:vlcache_method}


Motivated by observations from preliminary experiments, we introduce our method \vlcache, which strategically combines sparsity-aware cache budget allocation and modality-aware token scoring policy to improve VLM's performance under limited KV cache budget, in terms of both accuracy and efficiency.

\begin{figure}[h]
    \centering
    \includegraphics[width=0.8\textwidth]{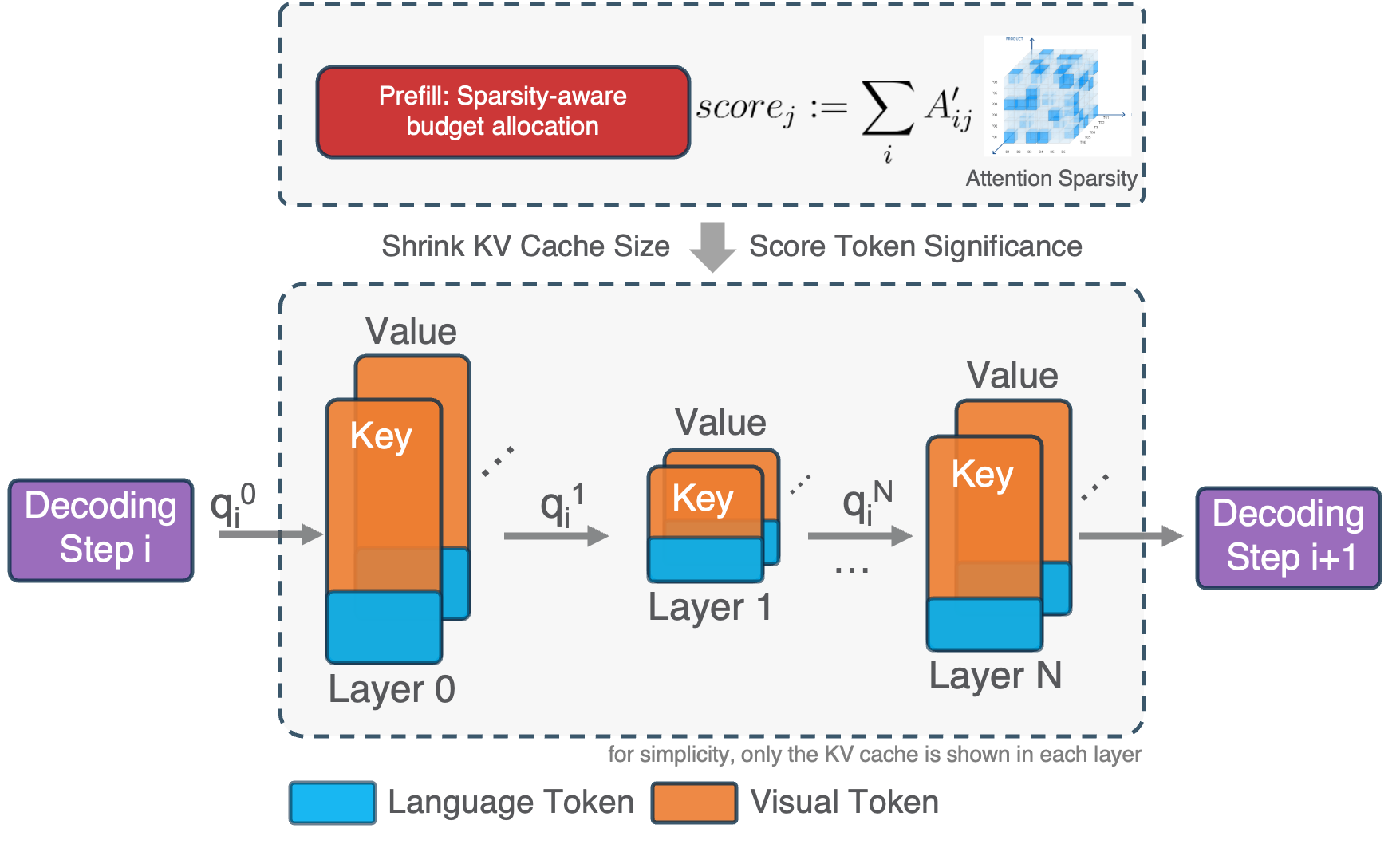}
    \caption{VL-Cache Overview. In the prefill stage, the cache budget for each layer is dynamically allocated according to the layer-wise sparsity. Then Post-vision Attention is employed to select both critical visual and language tokens.}
    \label{fig:arch}
\end{figure}

Specifically, we use Post-vision Attention to compute both inter-layer sparsity and intra-layer token importance. The former guides how many cache tokens should be allocated at each layer, while the latter dictates which $k$ tokens within a layer should be kept due to their importance. A high-level description of our method is visualized in Figure \ref{fig:arch}. For brevity, we will use $A'$ to denote the Post-vision Attention matrix in this section. 

We argue that using Post-vision Attention has two important advantages. Firstly, computing layer-wise sparsity using Post-vision Attention results in much lower memory and latency $O(\tau m)$ compared to using the full attention matrix $O(m^{2})$ as in \cite{zhang2024h2o}, since the visual tokens dominate the prompt length, $\tau \ll m$ for current vision-language tasks. Second, with a fixed cache budget, using Post-vision Attention leads to better preservation of important tokens, as measured by high $\CacheHitRate$ in Figure \ref{fig:cachehitrate}. In this section, we will detail how we perform sparsity-aware KV cache budget allocation and cache token eviction guided by Post-vision Attention before reviewing the strong performance of \vlcache\ in Section \ref{section:exp}.

\subsection{Sparsity-aware KV Cache Budget Allocation}
\label{kv_cache_budget_allocation}

Before determining the exact tokens to evict, we need to allocate the KV cache budget, which is the percentage of the KV cache to retain at each layer. Based on our observations from Section \ref{observation:layer-wise sparsity}, we implement a \textit{sparsity-aware layer-wise KV cache allocation} approach with two steps during the prefill phase. First, we apply $\ThresholdFilter$ pruning (with $p=1\%$) to the Post-vision Attention scores and calculate the layer-wise sparsity. Second, given a target KV cache budget for the whole model, we distribute this budget across layers based on each layer's sparsity ratio. This method optimizes the use of limited memory to store an appropriate amount of context information in each layer. Note that the allocation occurs only once and right after prefill phase, so the latency overhead can be amortized across multiple decoding steps.

\begin{algorithm}
\caption{Sparsity-Aware Cache Budget Allocation}
\label{alg:allocate_cache_budget}
\begin{algorithmic}[1]
\State \textbf{Input:} query and key $Q, K \in \mathbb{R}^{L \times H \times m \times d}$, number of layers $L$, number of heads $H$, length of post-vision prompt ${\tau}$, cache budget $\alpha$.
\State \textbf{Output:} layer-wise budget $\beta$
\Procedure{ComputeSparsity}{$Q, K$}
\State $Q' \gets Q_{m - \tau:m}$ 
\State $A' \gets softmax(\frac{Q'K^T}{\sqrt{d}})$

\State $\gamma' \gets \sum_{i + m - \tau \geq j}\frac{\mathbbm{1}\lbrack{\ThresholdFilter(A', p)_{ij} = 0}\rbrack}{{|\{A_{ij}': i + m - \tau \geq j\}|}}$
\State return $\gamma'$
\EndProcedure
\\
\Procedure{SkewedCacheBudgetAllocation}{$Q, K, \alpha, L, H$}
    
    \State $\Gamma[L][H] \gets 0$
    \For{$l = 1 \rightarrow L$}
        \For{$h = 1 \rightarrow H$}
            \State $\Gamma_{h}^{(l)} \gets \text{ComputeSparsity}(Q_{h}^{(l)}, K_{h}^{(l)})$ 
        \EndFor
    \EndFor

    \State $\gamma' \gets \Gamma . \text{mean}(1)$, $Z \gets \sum_{l}{1 -\gamma^{'(l)}} $, $\beta[L] \gets 0$
    \For{$l = 1 \rightarrow L$}
        \State $\beta^{(l)} \gets \text{clip(} \frac{1.0-\gamma^{'(l)}}
        {Z} \alpha L, 0.01, 1\text{)}$
    \EndFor

    \State \textbf{return} $\beta$
\EndProcedure

\end{algorithmic}
\end{algorithm}

In Algorithm \ref{alg:allocate_cache_budget}, we present our KV cache budget allocation algorithm, where $\alpha$ is a hyper-parameter that encodes the desired KV cache budget for the whole model. When the accuracy is not satisfactory, a higher $\alpha$ can be used to keep more KV cache. Finally, we would like to point out the key difference between our algorithm and PyramidKV \citep{zhang2024pyramidkv}: our allocation budget is customized for each prompt based on its sparsity pattern, instead of using a fixed layer-wise budget for all prompts, thus granting additional flexibility. 

\subsection{Modality-Aware Token Scoring Policy}

After we decide the cache budget for layer $l$, we choose a subset of $k^{(l)}$ cache tokens from the full cache. Prior works (that were targeting KV cache compression for LLMs) have explored several scoring policies, such as \textit{Accumulated Attention} in H2O \citep{zhang2024h2o} and \textit{Accumulated Sliding Window Attention} in \cite{li2024snapkv, zhang2024pyramidkv, yang2024pyramidinfer}, as we discussed earlier in Section \ref{sec: cachehitrate}. In this section, we will explain in-depth why \textit{Accumulated Post-vision Attention} provides a better scoring policy than prior policies for VLMs.

While \textit{Accumulated Attention} serves as a simple yet effective baseline, one critical issue is that summation over the full query dimension unavoidably assigns high scores to earlier tokens. Observing this length bias, \textit{Accumulated Sliding Window Attention} only sums attention scores over a recent window. We confirm that this indeed leads to higher $\CacheHitRate$, as shown in Figure \ref{fig:cachehitrate}. However, as we can see in Figure \ref{fig:llm_attn} and \ref{fig:vlm_attn}, VLMs have a significantly different attention pattern than LLMs. Along the query dimension, we observe a clear modality boundary: visual tokens attend to other visual tokens rather uniformly while language tokens only pay attention to a few visual tokens with high concentration. If we were to apply \textit{Accumulated Attention} as the scoring policy, the signal for critical cache tokens will be buried as we sum scores from all the preceding visual tokens. Using a recent window helps, but a fixed window size can easily be too large or too small as the length of the post-vision prompt varies from case to case. Therefore, we introduce \textit{Accumulated Post-vision Attention} as an optimized scoring policy for VLMs by implementing a dynamic, prompt-specific window size. The advantage of our scoring policy is evident by its high $\CacheHitRate$ in Figure \ref{fig:cachehitrate}, and will be further demonstrated through accuracy comparisons in Section \ref{section:exp}.

\section{Experiments}
\label{section:exp}
In our experiments, we evaluate \vlcache\ across representative VLMs that can handle image, video and language inputs. We use the state-of-the-art open-source LLaVA family, including LLaVA-Mistral-7B \citep{liu2023llava} and LLaVA-1.6-34B \citep{liu2023improvedllava}.
As shown in Table \ref{tab:llava_family}, they all share the same visual model (openai/clip-vit-large-patch14-336 \citep{radford2021learning}) but are fine-tuned from different language backbones (e.g., Mistral \citep{jiang2023mistral}, Nous-Hermes-2-Yi-34B \citep{noushermes2023}). Also, the former model uses Grouped Query Attention (GQA) \citep{ainslie2023gqa}, while the latter model uses Multi-Head Attention (MHA) \citep{vaswani2017attention}.

\textbf{Implementation Details}. We use an AWS EC2 P4 instance equipped with 8 A100 40GB GPUs for evaluation. 
First, we sample three tasks from lmms-eval \citep{lmms_eval2024}, including Coco-Caption \citep{chen2015microsoft}, DocVQA \citep{mathew2021docvqa}, and MathVista \citep{lu2023mathvista}. These tasks are representative, and cover OCR, reasoning, and world knowledge domains. Second, we compare accuracy of our approach against full-cache baselines and previous KV cache sparsification methods including StreamingLLM \citep{xiao2023efficient}, H2O \citep{zhang2024h2o} and PyramidKV \citep{zhang2024pyramidkv}. We apply KV cache sparsification in line with these baselines by retaining the most recent tokens and selecting the Top-K tokens according to their corresponding scoring policies. All baselines are configured with their default settings, except that the KV cache budget is scaled proportionally to the prompt length, and the recent token window size is fixed at 10\% of this budget to enable a fair comparison. Finally, we benchmark latency with varying context lengths and batch sizes. 

\begin{table}[htbp]
    \centering
    \begin{tabular}{>{\raggedright}p{3cm} >{\raggedright}p{3cm} >{\raggedright}p{3cm} >{\raggedright\arraybackslash}p{3cm}}
        \toprule
        \textbf{Model} & \textbf{Visual Model} & \textbf{Language Model} & \textbf{Attention Type} \\
        \midrule
        llava-v1.6-mistral-7b & openai/clip-vit-large-patch14-336 & Mistral-7B-Instruct-v0.2 & GQA \\
        llava-v1.6-34b & openai/clip-vit-large-patch14-336 & Nous-Hermes-2-Yi-34B & MHA \\
        \bottomrule
    \end{tabular}
    \caption{LLaVA Family Architecture}
    \label{tab:llava_family}
\end{table}




\subsection{Accuracy Evaluation}
\label{acc_eval}
\begin{figure}[h]
    \centering
    \begin{subfigure}[b]{0.32\textwidth}
        \label{coco-mistral-7b}
        \centering
        \includegraphics[width=\textwidth]{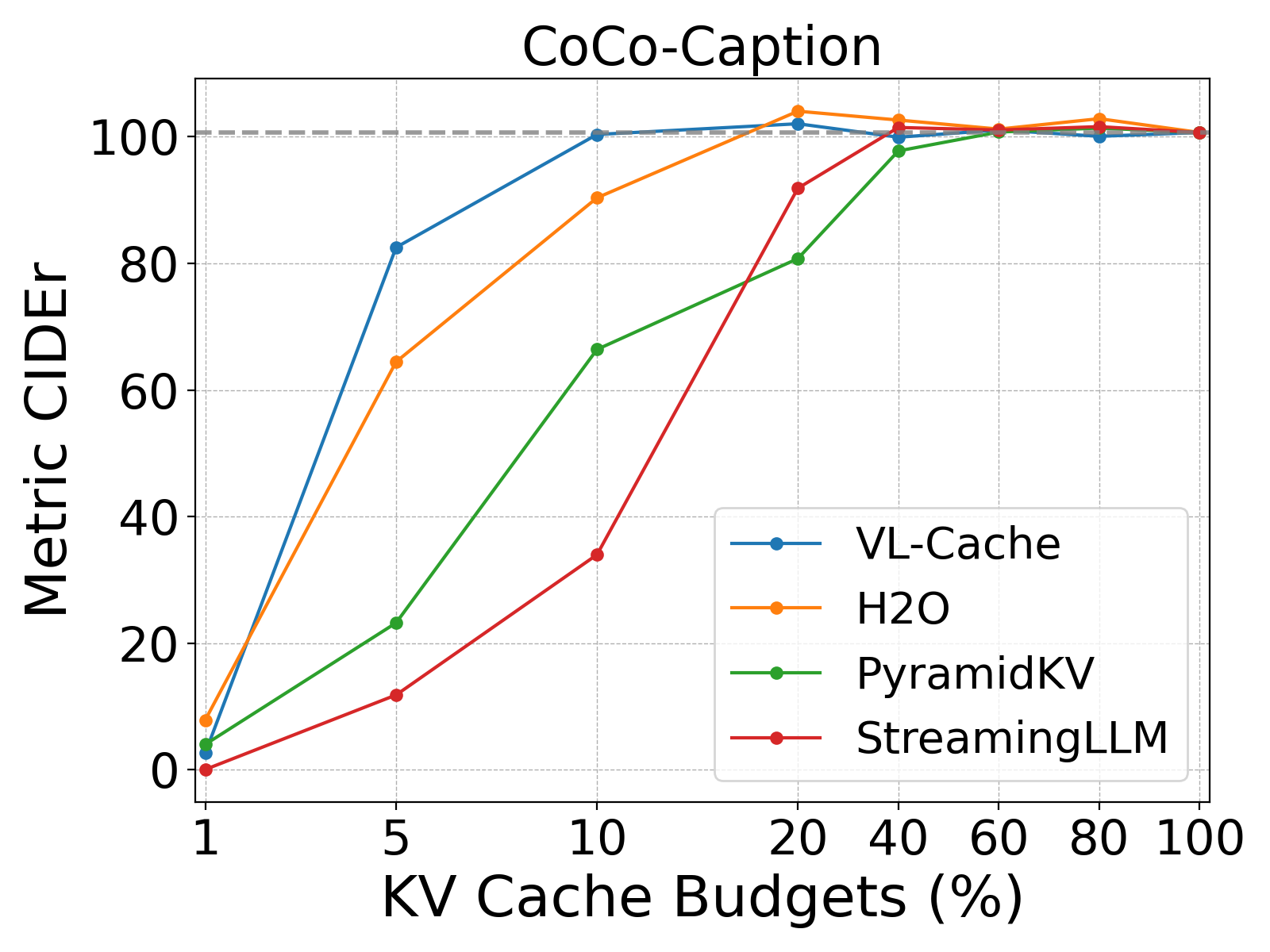}
        \caption{LLaVA-V1.6-Mistral-7B}
    \end{subfigure}
    \hfill
    \begin{subfigure}[b]{0.32\textwidth}
        \label{docvqa-mistral-7b}
        \centering
        \includegraphics[width=\textwidth]{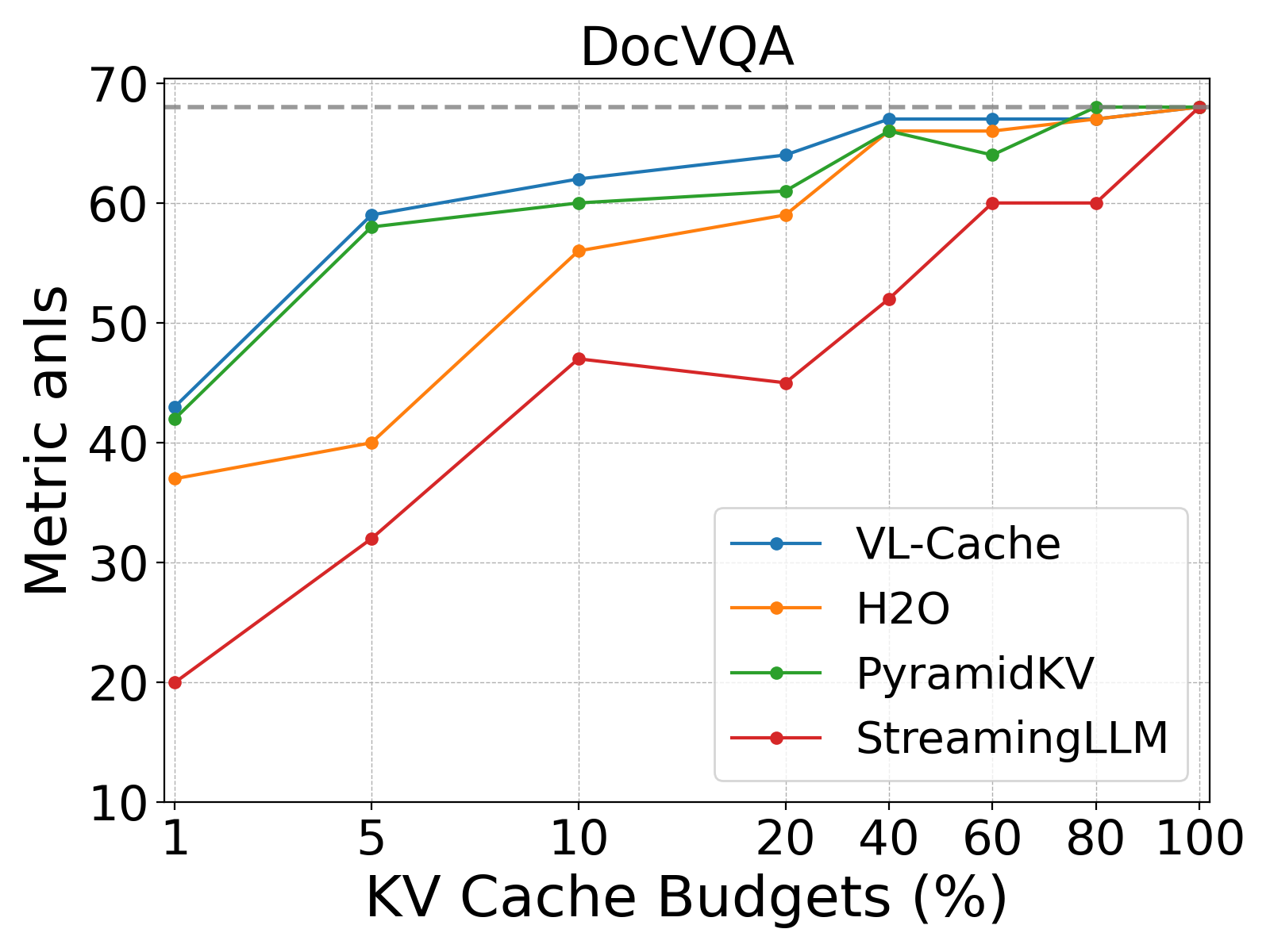}
        \caption{LLaVA-V1.6-Mistral-7B}
    \end{subfigure}
    \hfill
    \begin{subfigure}[b]{0.32\textwidth}
        \centering
        \includegraphics[width=\textwidth]{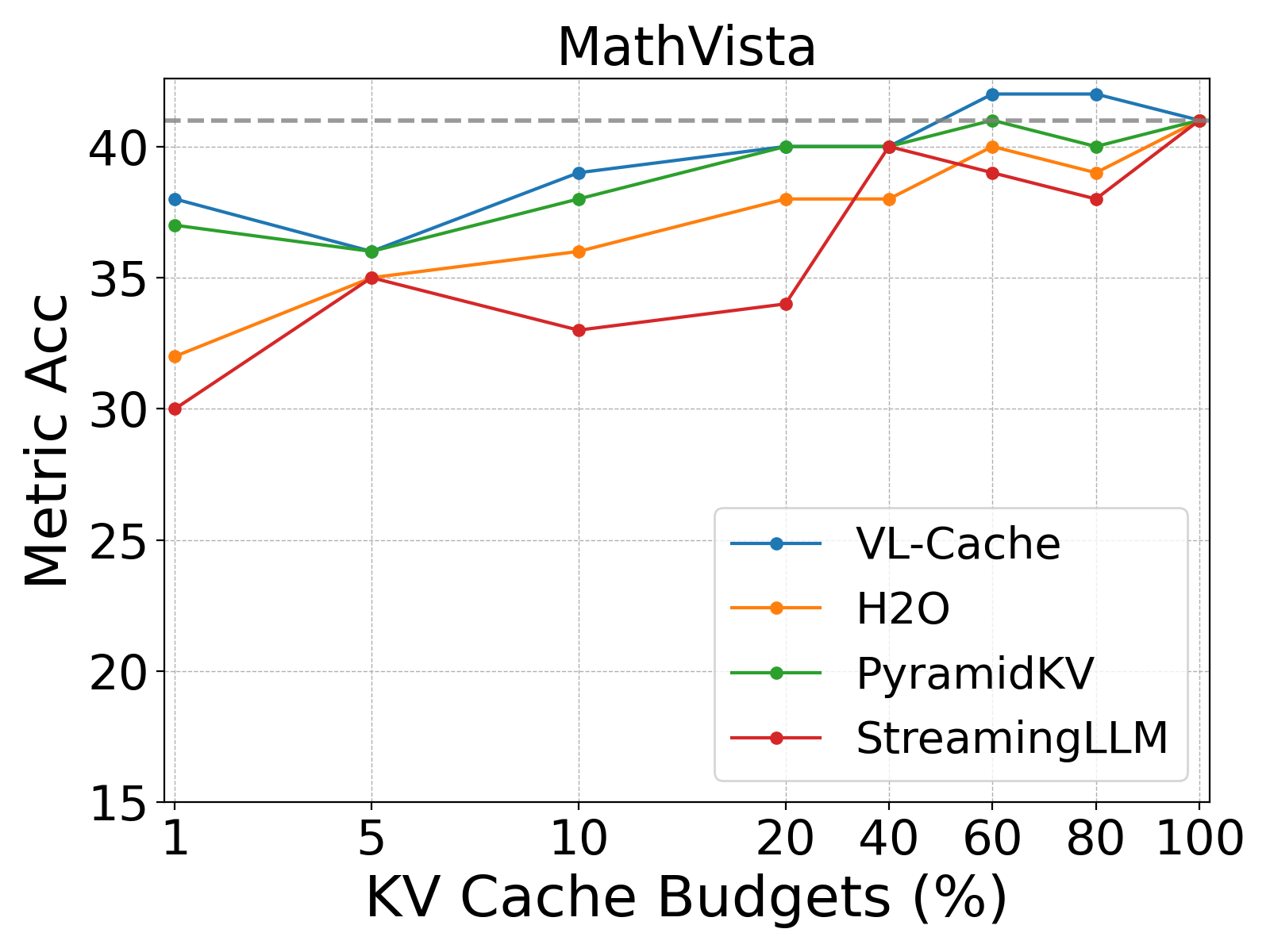}
        \caption{LLaVA-V1.6-Mistral-7B}
    \end{subfigure}

    \vskip\baselineskip
    \begin{subfigure}[b]{0.32\textwidth}
        \label{coco-llava-34b}
        \centering
        \includegraphics[width=\textwidth]{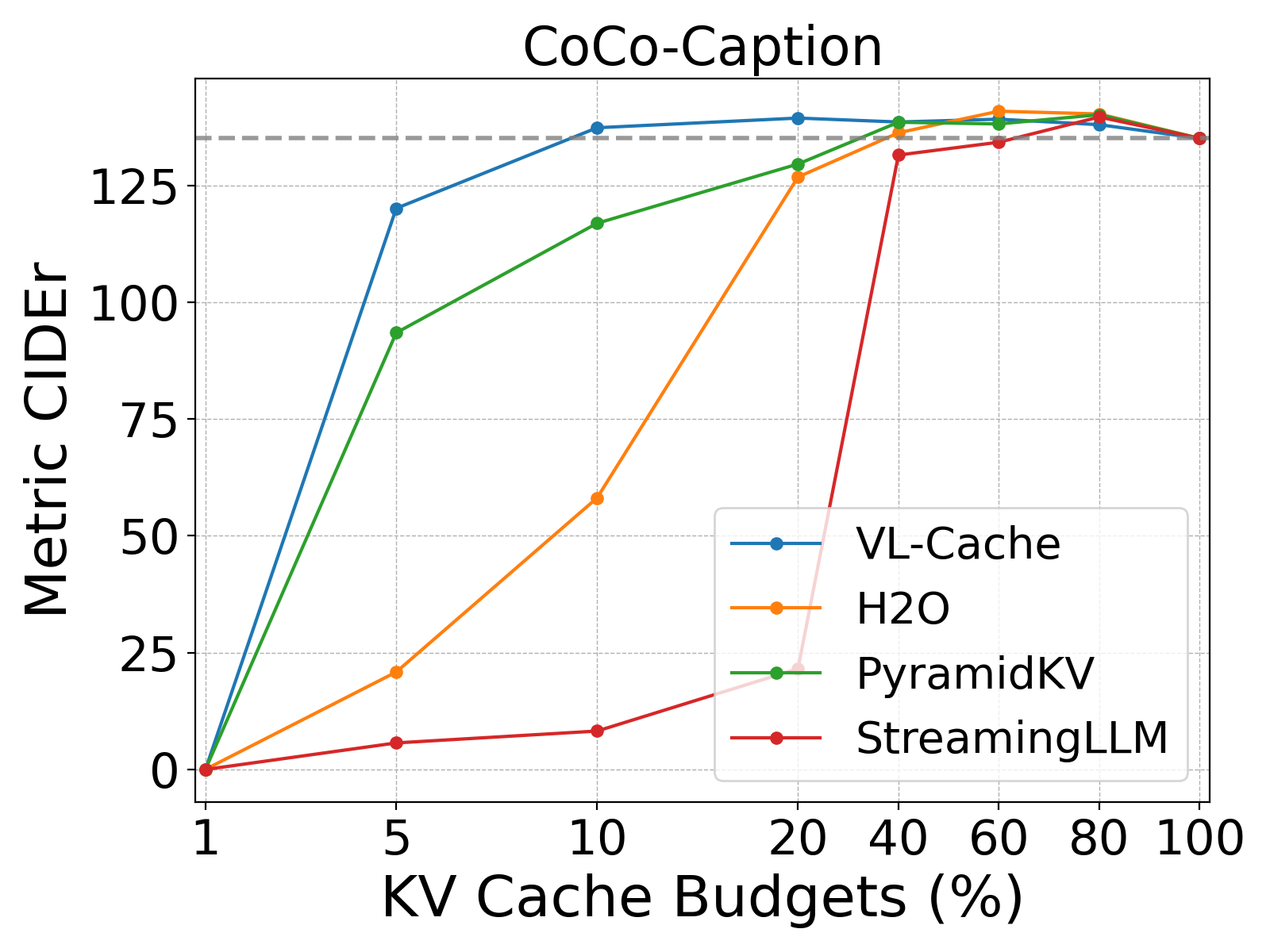}
        \caption{LLaVA-V1.6-34B }
    \end{subfigure}
    \hfill
    \begin{subfigure}[b]{0.32\textwidth}
        \label{docvqa-llava-34b}
        \centering
        \includegraphics[width=\textwidth]{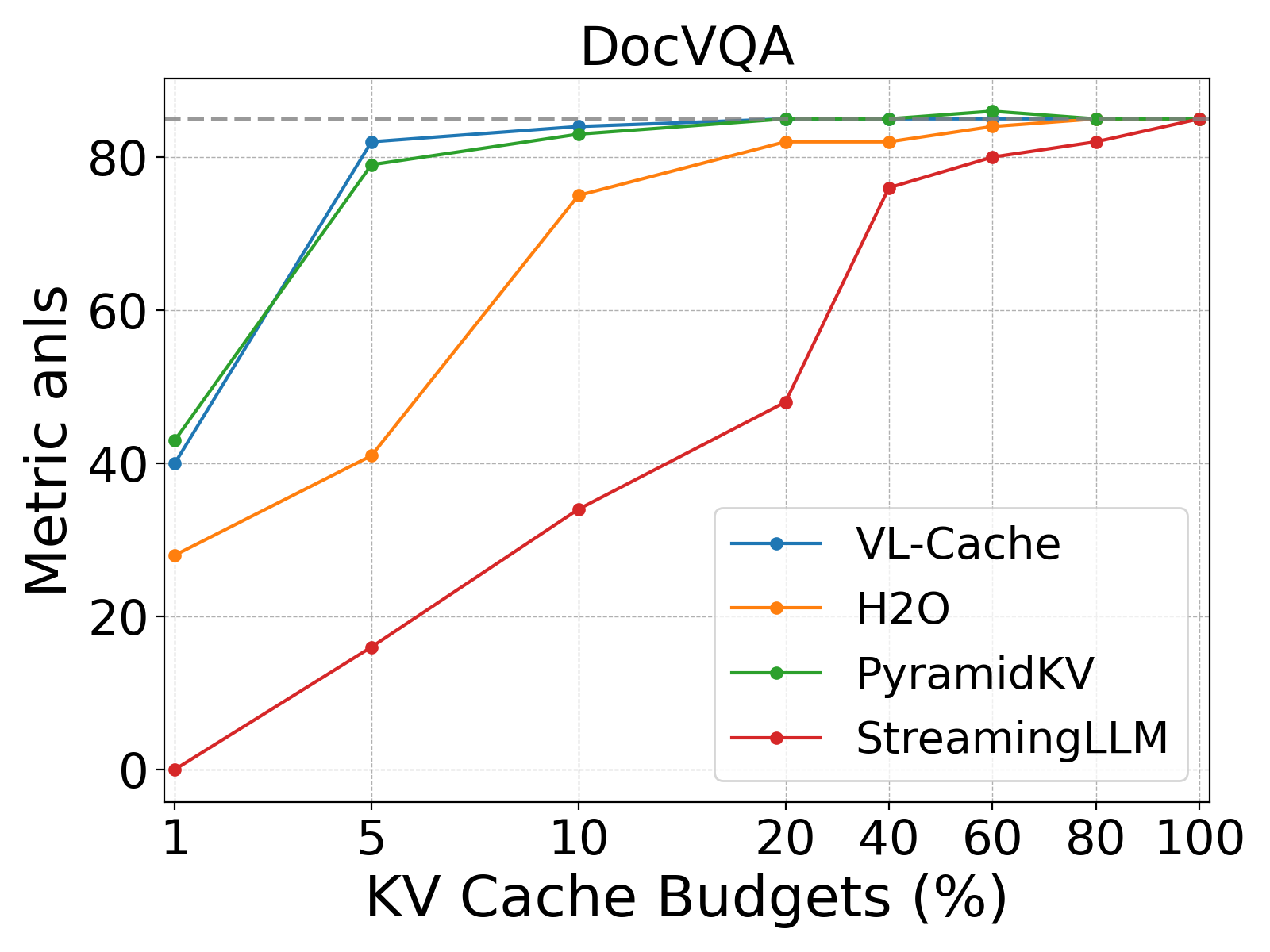}
        \caption{LLaVA-V1.6-34B}
    \end{subfigure}
    \hfill
    \begin{subfigure}[b]{0.32\textwidth}
        \centering
        \includegraphics[width=\textwidth]{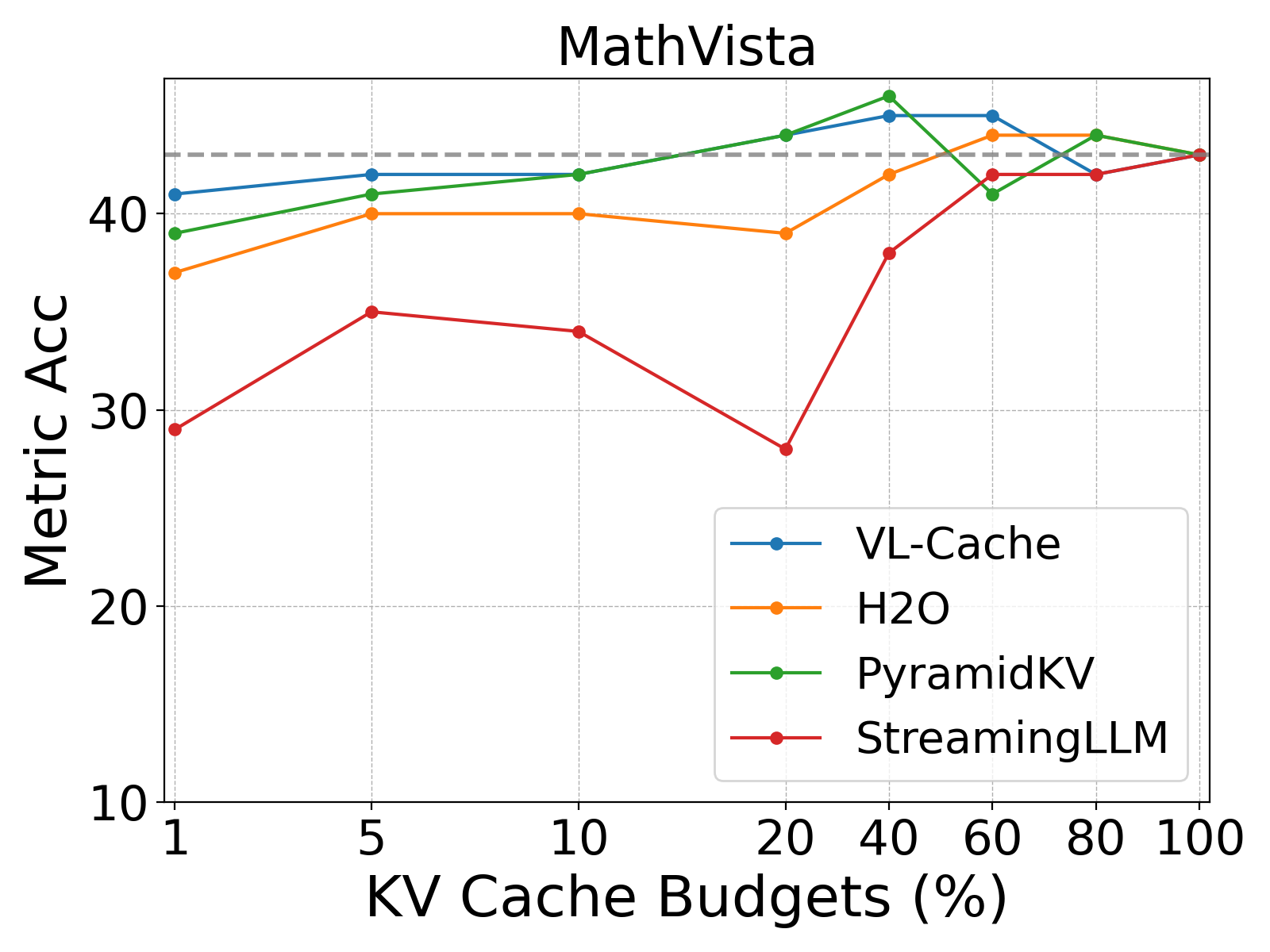}
        \caption{LLaVA-V1.6-34B}
    \end{subfigure}

    \caption{Evaluation results on different datasets with varied cache budgets. The evaluation metrics are the average of sampled tasks. \vlcache\ achieves comparable accuracy against full-cache and outperforms multiple baselines with limited KV cache budget.}
    \label{fig:accuracy}
    \vspace{-1em}
\end{figure}

The accuracy evaluation results are shown in Figure \ref{fig:accuracy} and Table \ref{tab:accuracy}. We report the average accuracy score with KV cache budget varying from 1\% to 100\% of prompt length. Overall, \vlcache\ outperforms other compression methods across a range of KV cache budgets and different language model backbones. 

For the Coco-Caption dataset (Figure \ref{fig:accuracy} (a), (c) and Table \ref{tab:accuracy}), all baselines maintain high accuracy when the KV cache budget exceeds 40\% for VLMs. However, as the KV cache budget is further reduced, accuracy significantly declines. It is worth noting that even with only 5\% to 10\% KV cache budget, \vlcache\ can achieve accuracy that is comparable to the full KV cache. H2O and StreamingLLM allocate the same cache budget across all layers, causing denser layers to miss important tokens in the context, while sparser layers contain redundant tokens. PyramidKV statically allocates KV cache in a monotonically decreasing manner, which is ineffective for all input queries.

For the DocVQA dataset (Figure \ref{fig:accuracy} (b), (d) and Table \ref{tab:accuracy}), PyramidKV demonstrates strong performance; however, our \vlcache\ consistently outperforms all other baselines. H2O and StreamingLLM lack robustness, with performance significantly declining as the KV cache is further compressed.

\begin{table}[ht]
\centering
\resizebox{\textwidth}{!}{%
\begin{tabular}{l|l|l|cccccccc}
\toprule
\textbf{Dataset} & \textbf{Model} & \textbf{Method} & \textbf{1\%} & \textbf{5\%} & \textbf{10\%} & \textbf{20\%} & \textbf{40\%} & \textbf{60\%} & \textbf{80\%} & \textbf{100\%} \\ 
\midrule
\multirow{8}{*}{\shortstack{Coco-Caption \\ (Metric: CIDEr)}} 
                 & \multirow{4}{*}{LLaVA-Mistral-7B}  & VL-Cache    & 2.64           & \textbf{82.53} & \textbf{100.36} & 102.06          & 99.93           & 101.07          & 100.08           &100.68 \\
                 &                                    & H2O         & \textbf{7.87}  & 64.45          & 90.36           & \textbf{104.04} & \textbf{102.64} & \textbf{101.21} & \textbf{102.86}  & 100.68 \\
                 &                                    & PyramidKV   & 4.01           & 23.21          & 66.41           & 80.76           & 97.76           & 100.75 & 101.38                    & 100.68 \\
                 &                                    & StreamingLLM& 0.05           & 11.82          & 33.98           & 91.87           & 101.47          & 101.07 & 101.6                     & 100.68 \\ 
\cmidrule{2-11}
                 & \multirow{4}{*}{LLaVA-1.6-34B}  & VL-Cache    & 0              & \textbf{120.11} & \textbf{137.35} & \textbf{139.42} & \textbf{138.58} & 139.19          & 138.01          & 135.07 \\
                 &                                 & H2O         & 0  & 20.87           & 58.14           & 126.8           & 136.27          & \textbf{140.89} & \textbf{140.3}  & 135.07 \\
                 &                                 & PyramidKV   & 0              & 93.47           & 116.91          & 129.59          & 138.53          & 138.17          & 140.15          & 135.07 \\
                 &                                 & StreamingLLM& 0              & 5.69            & 8.23            & 21.57           & 131.51          & 134.26          & 139.65          & 135.07 \\
\midrule
\multirow{8}{*}{\shortstack{DocVQA \\ (Metric: ANLS)} }     
                 & \multirow{4}{*}{LLaVA-Mistral-7B} & VL-Cache    & \textbf{43}    & \textbf{59}    & \textbf{62}     & \textbf{64}    & \textbf{67}   & \textbf{67}   & 67         & 68     \\
                 &                                   & H2O         & 37             & 40             & 56              & 59             & 66            & 66            & 67         & 68     \\
                 &                                   & PyramidKV   & 42             & 58             & 60              & 61             & 66            & 64            & \textbf{68}& 68     \\
                 &                                   & StreamingLLM& 20             & 32             & 47              & 45             & 52            & 60            & 60         & 68     \\ 
\cmidrule{2-11}
                 & \multirow{4}{*}{LLaVA-1.6-34B}     & VL-Cache    & 41            & \textbf{82}    & \textbf{84}     & \textbf{85}          & \textbf{85}     & 85             & 85     & 85     \\
                 &                                    & H2O         & 28            & 41             & 75              & 82          & 82              & 84             & \textbf{85}     & 85     \\
                 &                                    & PyramidKV   & \textbf{43}   & 79             & 83              & \textbf{85} & \textbf{85}     & \textbf{86}    & \textbf{85}     & 85     \\
                 &                                    & StreamingLLM& 0             & 16             & 34              & 48          & 76              & 80             & 82              & 85     \\ 
\midrule
\multirow{8}{*}{\shortstack{MathVista\\(Metric: ACC)}}   
                 & \multirow{4}{*}{LLaVA-Mistral-7B} & VL-Cache    & \textbf{38}    & \textbf{36}    & \textbf{39}      & \textbf{40}     & \textbf{40}     & \textbf{42}     & \textbf{42}     & 41     \\
                 &                                   & H2O         & 32             & 35             & 36               & 38              & 38              & 40              & 39              & 41     \\
                 &                                   & PyramidKV   & 37             & \textbf{36}    & 38               & \textbf{40}     & \textbf{40}     & 41              & 40              & 41     \\
                 &                                   & StreamingLLM& 30             & 35             & 33               & 34              & \textbf{40}     & 39              & 38              & 41     \\ 
\cmidrule{2-11}
                 & \multirow{4}{*}{LLaVA-1.6-34B}    & VL-Cache    & \textbf{41}	& \textbf{42}	  & \textbf{42}	   &\textbf{44}	   & 45            &\textbf{45}   & 42           & 43    \\
                 &                                   & H2O         & 37	        & 40	          & 40	           & 39	           & 42	           & 44	          & \textbf{44}  & 43               \\
                 &                                   & PyramidKV   & 39	        & 41	          & \textbf{42}	   & \textbf{44}   & \textbf{46}   & 41	          & \textbf{44}	 & 43              \\
                 &                                   & StreamingLLM& 29	        & 35	          & 34	           & 28	           & 38	           & 42	          & 42	         & 43              \\ 
\bottomrule
\end{tabular}
}
\caption{Performance of VLMs with different compression methods and datasets}
\label{tab:accuracy}
\end{table}








\subsection{Speed Benchmark}

In order to show the speed advantage of \vlcache, we measure the GPU kernel latencies of prefill and decoding forward passes with synthetic prompts, following the method in \cite{kwon2023efficient}. We vary the size of the prompt from 1K tokens to 128K tokens to scale our method to a very long context. Batch sizes vary from 1 to 64 and are static, meaning that all requests get prefilled and decoded concurrently. 
We assume that the prompt template format remains similar to our accuracy benchmarks, so we use the last 50 tokens of the prompt to determine which tokens to evict from the KV cache. For both prefill and decoding in the baseline, we used default settings from the HuggingFace implementation \footnote{https://huggingface.co/docs/transformers/main/en/perf\_infer\_gpu\_one}, including CUDA-based FlashAttention-v2. To optimize performance in our VL-Cache, we applied our Triton-based solution for self-attention forward pass, layer-wise sparsity evaluation, and modality-aware token scoring, as detailed in the appendix \ref{appendix:efficient_implementation}. The speedup is calculated as $\frac{\textbf{Baseline latency}}{\textbf{VLCache latency}}$.

In Table \ref{table:performance_metrics}, we observe that with 50 query tokens for calculating attention statistics, the overhead of our method is just 1-6\% of the prefill latency. See Appendix \ref{appendix_speed_benchmark} for detailed measurements of the overhead and a discussion on how to reduce the overhead of statistics calculation for a large number of query tokens. During decoding, we run 99 forward passes for a total of 100 output tokens. We observe up to 7x decoding speedups that are attributed to the reduced size of the KV cache, which we compressed to 10\% for this benchmark.

Overall, we see that the end-to-end speedup is bounded by the prefill latency, which is mostly unchanged by \vlcache. For example, with prompt length of 128K and batch size of 1, the decoding speedup of 7.08x is diluted by prefill taking 53\% of end-to-end latency, which results in 1.66x end-to-end speedup. End-to-end speedup will monotonically approach the decoding speedup as the count of output tokens increases. This highlights that KV cache sparsity will give the best speed advantage in tasks with long outputs, such as image captions, video descriptions, chain-of-thought multi-modal reasoning, etc.

\noindent 
\begin{minipage}{0.59\textwidth} 
    \centering
    \resizebox{\textwidth}{!}{
        \begin{tabular}{p{1cm} p{1.5cm} p{1.5cm} p{1.5cm} p{2.5cm}} 
            \toprule
            Batch Size & Prompt Length & Prefill Speedup & Decoding Speedup & End-to-End Speedup \\
            \midrule
            1 & 2000 & 0.96 & 1.19 & 1.16 \\
            1 & 8000 & 0.97 & 1.70 & 1.49 \\
            1 & 32000 & 0.99 & 3.32 & 1.85 \\
            1 & 128000 & 0.99 & 7.08 & 1.66 \\
            \midrule
            4 & 2000 & 0.98 & 1.68 & 1.50 \\
            4 & 8000 & 0.98 & 3.16 & 1.95 \\
            4 & 32000 & 0.99 & 6.07 & 2.06 \\
            \midrule
            16 & 2000 & 0.98 & 3.03 & 1.99 \\
            16 & 8000 & 0.99 & 5.61 & 2.27 \\
            \midrule
            64 & 2000 & 0.98 & 5.23 & 2.33 \\
            \bottomrule
        \end{tabular}
    }
    \captionof{table}{Performance metrics by batch size and prompt length for 100 output tokens.}
    \label{table:performance_metrics}
\end{minipage}
\hfill
\begin{minipage}{0.40\textwidth}
    \centering
    \includegraphics[width=\textwidth]{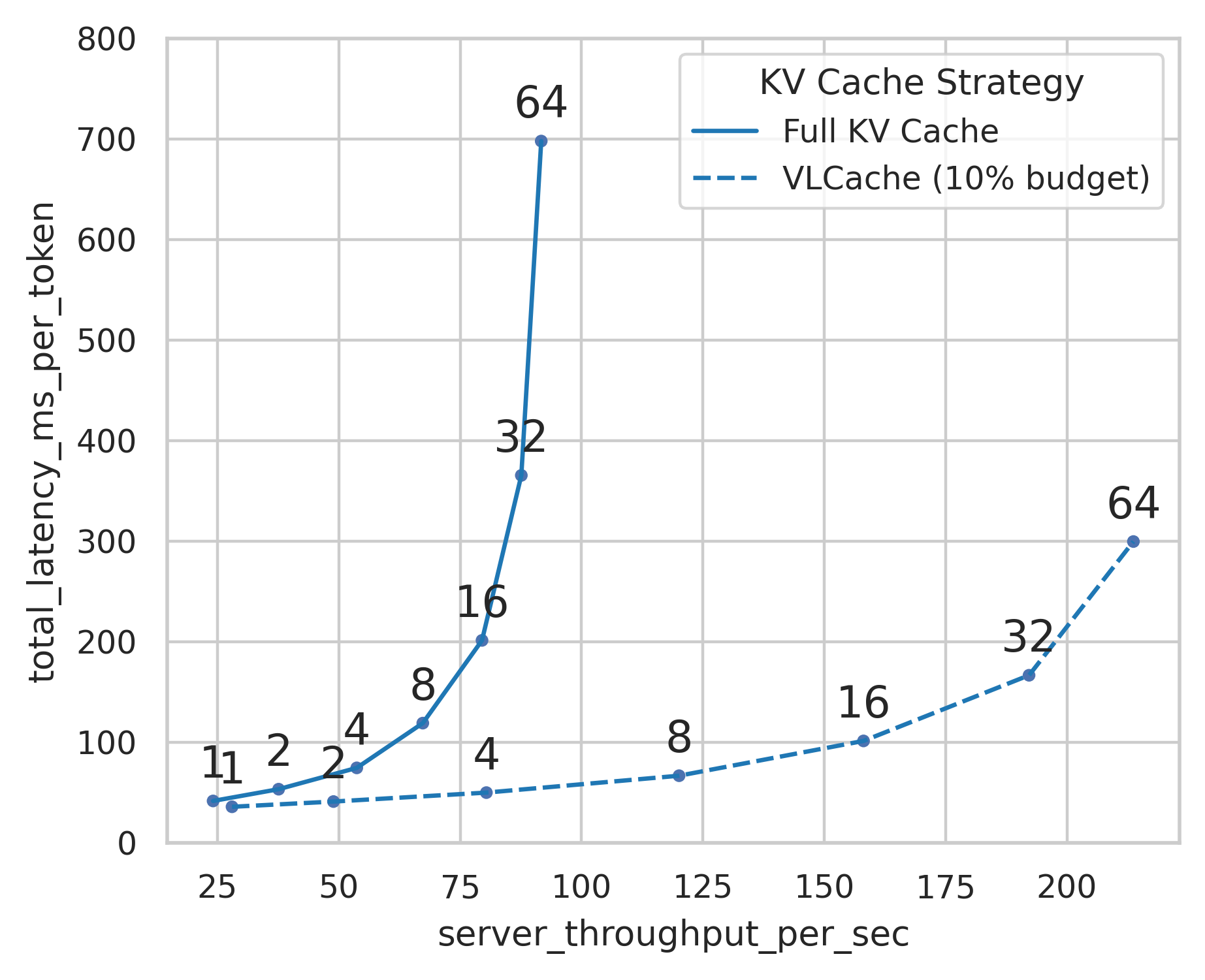}
    \vspace{-1em}
    \captionof{figure}{Server-level throughput v.s. request-level latency curve (prompt length = 2K). Labeled points indicate batch size.}
    \label{fig:latency_throughput_curve}
\end{minipage}


In our implementation of both the baseline and \vlcache, maximum batch size is limited by peak memory usage during prefill instead of KV cache size, so compression of KV cache does not lead to higher batch size. In future work, continuous batching and chunked prefill could be used to eliminate the prefill memory bottleneck, which would expose our method's advantage in raising maximum batch size when the size of KV cache is the bottleneck to higher batch size.

Finally, we summarize the trade-off between request-level latency and server-level throughput in Figure \ref{fig:latency_throughput_curve}. We note that \vlcache\ offers both higher peak throughput and lower latency for any desired server-level throughput.

\section{Conclusion}
In this paper, we propose VL-Cache, a novel KV cache compression optimized for VLMs. We discovered the unique sparsity patterns of visual and language tokens throughout the prefill and decoding phases. With these observations, we introduce a modality-aware token scoring policy and sparsity-aware cache budget allocation to reduce KV cache size without accuracy loss. Empirical results on multiple benchmark datasets demonstrate that when maintaining only 10\% of the KV cache, our method achieves accuracy comparable to the full KV cache and outperforms all existing methods. In a speed benchmark, our method accelerates end-to-end latency of generating 100 tokens by up to 2.33x relative to the full KV cache.




\bibliography{iclr2025_conference}
\bibliographystyle{iclr2025_conference}

\appendix

\section{Appendix}



\subsection{Vision-Language Prompt Template Construction}

\begin{figure}[h]
    \centering
    \includegraphics[width=\linewidth]{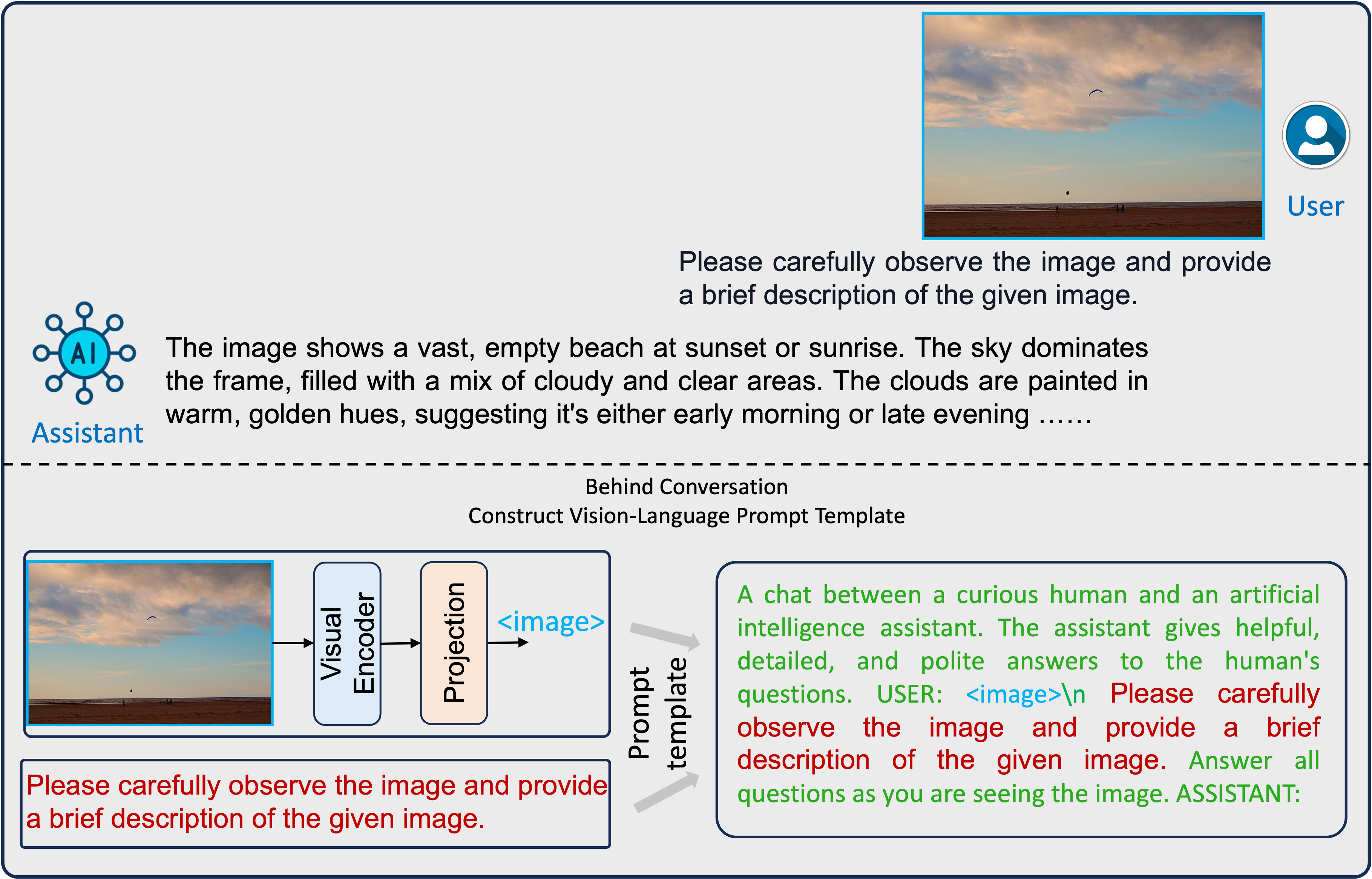}
    \caption{Vision-Language Prompt Template Construction. Regardless of the image order (before or after language inputs), with an appropriate prompt template design, there will always be a language-based instruction or question in the post-vision position. }
    \label{fig:prompt_template}
\end{figure}

Constructing prompt templates in image-based conversations is a common practice for VLMs \citep{li2024llava-onevision, team2023gemini, islam2024gpt, bai2023qwen, anthropic2023claude}, as it instructs language models to generate more accurate responses. For example, as illustrated in Figure \ref{fig:prompt_template}, the input image is processed through a visual encoder and a projection layer to generate an image embedding, represented by the \texttt{<image>}. For language input, beyond user input, a prompt template is employed. With the appropriate prompt template design, regardless of the original image order from user input (whether before or after language inputs), there will always be a language-based instruction or question in the post-vision position, providing a strong signal for our \vlcache\ to evict insignificant visual tokens.

\subsection{Extended Related Works}

The KV cache, while essential for transformer-based LLM and VLM family, demands significant computational resources and memory, limiting inference speed. To address these challenges, researchers have explored various KV cache compression techniques, such as KV cache sparsification, quantization, or a combination of both.

\textbf{KV cache sparsification}. Heavy-Hitters (H2O) \citep{zhang2024h2o} employs cumulative attention scores to greedily evict unimportant tokens. However, this method tends to accumulate more attention on the initial tokens, introducing bias and negatively impacting the identification of key tokens during decoding. ZipCache \citep{he2024zipcache} further normalizes the cumulative attention scores, leading to more precise prediction. Keyformer \citep{adnan2024keyformer} proposes a novel score function to predict the importance of each token and only window tokens and dynamic key tokens are kept in the KV cache. PyramidKV \citep{yang2024pyramidinfer} allocates progressively smaller KV cache sizes as layers get deeper. This approach can achieve an absolute accuracy improvement of nearly 20.5 on specific tasks.


\textbf{KV cache quantization}. This method reduces the size of the KV cache by utilizing lower-bit representations of the key and value pairs. In particular, FP8 and INT8 quantization of KV caches are commonly used for LLM inference. Further compression of KV cache into lower-bits is also being explored, e.g. KIVI \citep{liu2024kivi} compresses KV cache into 2-bit representations.

\textbf{Combination of quantization and sparsification}. Quantization and sparsification \citep{yang2024no} technically could have orthogonal implementations that can be enabled together to obtain further. However, since both are lossy methods that can impact the accuracy of downstream tasks, careful accuracy evaluation of the quality of the output need to be conducted. 

The methods mentioned above mostly address LLM only and there is limited exploration for KV cache sparsification in VLMs. \vlcache\ proposed in this paper analyzes and utilizes the unique sparsity pattern in VLMs, which results in better accuracy for VLM inference.

\subsection{Efficient Implementation}
\label{appendix:efficient_implementation}
In order to discard tokens that received low attention during prefill, we need to calculate 2 statistics from the attention score matrix during the prefill stage: (1) the average attention score for each token in key dimension; and (2) the count of attention scores that are smaller than $p\%$ of the maximum within the query $Q$ dimension. 

Regular attention kernels, such as FlashAttention \citep{dao2023flashattention} and PagedAttention \citep{kwon2023efficient}, do not materialize the attention scores in HBM, so we need a new memory-efficient kernel to calculate the attention statistics. Overall, we aim to schedule these operations to ensure that: (1) The attention mask is not written to HBM; (2) the $QK$ product is not written to HBM; (3) the attention scores are not written to HBM; (4) similarly-parallelized operations (e.g. div and add) are fused to avoid intermediate tensors in HBM.

In the initial attempt, we used TorchCompiler with the Triton backend, which addressed requirements (1), (3), and (4), but not (2). An efficient fusion of the matmul operation with softmax is described in the FlashAttention paper and can be implemented in Triton, but TorchCompiler was not able to apply this optimization. We will now describe the optimal algorithm that satisfies all performance requirements.

The key to efficient fusion of matmul and softmax is to partition the work along the $Q$ dimension and then hold a single tile of $Q$ tensor in SRAM, while continuously loading tiles of the other tensor and computing the reductions without writing the $QK$ product back to HBM. The 2 reductions required by softmax (pre-exponentiation max, and post-exponentiation sum) are both along the $K$ dimension, so partitioning along the $Q$ dimension is required for these operations. The online softmax algorithm can be used to update the partially computed sum.

For average attention score in the KV context, we need to reduce the attention scores over the $Q$ dimension, which is not aligned with the previous computation. Instead, we need a second kernel which partitions the work over the $K$ dimension, so that a compute block can reduce over the $Q$ dimension. However, since we do not want to materialize the attention scores, the second kernel needs to recompute softmax again. In order do that, we re-use the max and sum over Q dimension from the first kernel, and these 2 tensors (both $O(Q)$ space) are the only intermediate tensors that are stored in HBM.

For sparsity ratio, counting the scores below the threshold is not possible until the maximum along the $K$ dimension is known. We can use the intermediate maximum tensor that was produced by the first kernel to count the sparse scores in the second kernel. Since each thread block will output its own count, we need the 3rd kernel to sum up the partial counts.

In summary, the optimized computation consists of 3 kernels executed serially:

\begin{enumerate} 
\item Softmax + row-wise stats (reductions over K dimension). Reads: $Q$ and $K$ from HBM. Computes and writes to HBM:
\begin{enumerate}
    \item pre-exp max over $K$ dimension (used in softmax for numerical stability)
    \item post-exp sum over $K$ dimension (used in softmax as the normalization factor)
\end{enumerate}
\item Softmax + column-wise stats (reductions over Q dimension). Reads: Q, K, pre-exp max, and post-exp sum from HBM. Computes and writes to HBM:
\begin{enumerate}
    \item the post-softmax sum over the Q dimension
    \item count of attention scores below threshold over the Q dimension
\end{enumerate}
\item Sum reduction of the partial count of attention scores below threshold
\end{enumerate}

We implemented these 2 kernels in Triton and observed that each had roughly the same latency as the FlashAttention kernel, implying that this partioning is optimal. However, even with this approach, we would want to use only a few dozen tokens for querying statistics during prefill.

\subsection{Speed Benchmark Results}
\label{appendix_speed_benchmark}

In this section, we report detailed metrics from the speed benchmark results. We will also describe the speed benchmark methodology a little more. We used Torch Profiler to measure GPU kernel latencies, and then added up the latencies from different kernels to get a total latency of a particular operation. This breakdown of latency by GPU kernels enabled us to 1) avoid measuring CPU latency that does not relate to our method, 2) group the latency into different buckets for reporting. Empty cells mean that the 
 inference server crashed due to insufficient GPU memory during prefill.

The following table presents the cumulative latency of all GPU kernels that are related to computing attention statistics from 50 query tokens, eviction scores for all prompt tokens, and copying the KV cache tensor after eviction of 90\% of the KV cache into contiguous memory. This prefill overhead is stated in milliseconds:

\begin{tabular}{rrrrrrrrr}
\toprule
batch size & 1000 & 2000 & 4000 & 8000 & 16000 & 32000 & 64000 & 128000 \\
\midrule
1 & 5.20 & 7.20 & 10.10 & 17.30 & 28.30 & 49.70 & 100.20 & 189.60 \\
2 & 6.40 & 9.50 & 15.80 & 29.00 & 51.80 & 92.50 & 188.50 &   \\
4 & 8.50 & 15.20 & 27.40 & 52.50 & 93.30 & 175.80 &   &   \\
8 & 13.70 & 26.80 & 50.90 & 96.60 & 178.40 &   &   &   \\
16 & 25.20 & 51.30 & 95.60 & 183.50 &   &   &   &   \\
32 & 48.50 & 97.10 & 182.60 &   &   &   &   &   \\
64 & 91.60 & 186.70 &   &   &   &   &   &   \\
\bottomrule
\end{tabular}

We define the prefill latency as the latency of generating 1 output token -- starting with input token IDs and ending with output logits, not including tokenization and sampling. The following table presents the prefill speedup ($\frac{\textbf{prefill} + \textbf{overhead}}{\textbf{prefill}}$). We observe that the overhead is never more than 6\% and becomes smaller as the prompt gets longer:

\begin{tabular}{rrrrrrrrr}
\toprule
batch size & 1000 & 2000 & 4000 & 8000 & 16000 & 32000 & 64000 & 128000 \\
\midrule
1 & 0.94 & 0.96 & 0.97 & 0.97 & 0.99 & 0.99 & 0.99 & 0.99 \\
2 & 0.96 & 0.97 & 0.97 & 0.98 & 0.98 & 0.99 & 0.99 &   \\
4 & 0.97 & 0.98 & 0.98 & 0.98 & 0.98 & 0.99 &   &   \\
8 & 0.98 & 0.98 & 0.98 & 0.99 & 0.99 &   &   &   \\
16 & 0.98 & 0.98 & 0.99 & 0.99 &   &   &   &   \\
32 & 0.98 & 0.98 & 0.98 &   &   &   &   &   \\
64 & 0.99 & 0.98 &   &   &   &   &   &   \\
\bottomrule
\end{tabular}

We calculated decoding latency as the difference between end-to-end latency and the prefill latency. Since we use static batching with a constant number of output tokens, it's easy to decompose the end-to-end latency into its components without concerns for queue waiting, scheduler stalling, sequence termination, etc. The following table presents the decoding speedup ($\frac{\textbf{decoding latency with full KV cache}}{\textbf{decoding latency with 10\% of KV cache}}$):

\begin{tabular}{rrrrrrrrr}
\toprule
batch size & 1000 & 2000 & 4000 & 8000 & 16000 & 32000 & 64000 & 128000 \\
\midrule
1 & 1.10 & 1.19 & 1.37 & 1.70 & 2.31 & 3.32 & 5.17 & 7.08 \\
2 & 1.18 & 1.36 & 1.68 & 2.27 & 3.16 & 4.56 & 6.78 &   \\
4 & 1.36 & 1.68 & 2.26 & 3.16 & 4.39 & 6.07 &   &   \\
8 & 1.73 & 2.29 & 3.13 & 4.37 & 5.87 &   &   &   \\
16 & 2.19 & 3.03 & 4.18 & 5.61 &   &   &   &   \\
32 & 2.94 & 4.12 & 5.53 &   &   &   &   &   \\
64 & 3.83 & 5.23 &   &   &   &   &   &   \\
\bottomrule
\end{tabular}

The end-to-end speedup is calculated as $\frac{\textbf{Baseline latency}}{\textbf{VLCache latency}}$. VLCache latency includes prefill latency, VLCache overhead, and 99 decoding passes to generate a total of 100 output tokens:

\begin{tabular}{rrrrrrrrr}
\toprule
batch size & 1000 & 2000 & 4000 & 8000 & 16000 & 32000 & 64000 & 128000 \\
\midrule
1 & 1.09 & 1.16 & 1.30 & 1.49 & 1.71 & 1.85 & 1.89 & 1.66 \\
2 & 1.16 & 1.30 & 1.49 & 1.73 & 1.89 & 1.97 & 1.94 &   \\
4 & 1.30 & 1.50 & 1.74 & 1.95 & 2.06 & 2.06 &   &   \\
8 & 1.54 & 1.79 & 2.00 & 2.17 & 2.23 &   &   &   \\
16 & 1.75 & 1.99 & 2.17 & 2.27 &   &   &   &   \\
32 & 1.98 & 2.19 & 2.32 &   &   &   &   &   \\
64 & 2.18 & 2.33 &   &   &   &   &   &   \\
\bottomrule
\end{tabular}

\subsection{Measuring attention to Visual and Language Tokens}

As we have established in the Section \ref{observation:layer-wise sparsity}, the attention sparsity during decoding can be predicted from attention scores in prefill phase. To further understand the importance of visual tokens and language tokens in the prompt, we examine the division of attention from decoding tokens to these two modalities of prompt tokens. 

 We propose $\Contribution$ for quantitative analysis and visualization of the \textit{layer-wise} \textit{modality-specific} attention patterns. Let $T$ be the current sequence length, $t$ be the index of the first decoded token, and $A^{(l)} \in \mathbb{R}^{T \times T}$ be the attention matrix at layer $l$ for one specific head. 

\begin{equation}
    \Contribution_{mod}^{(l)} \coloneq \frac{1}{T-t+1} \sum_{i \geq t}{\frac{\sum_{j \in J_{mod}}{\ThresholdFilter(A^{(l)}, p)_{ij}}}{\sum_{j\in J_{all}}{\ThresholdFilter(A^{(l)}, p)_{ij}}}}
\end{equation}

\begin{figure}[h]
    \centering
    \begin{minipage}[t]{0.48\textwidth}
        \centering
        \includegraphics[width=\textwidth]{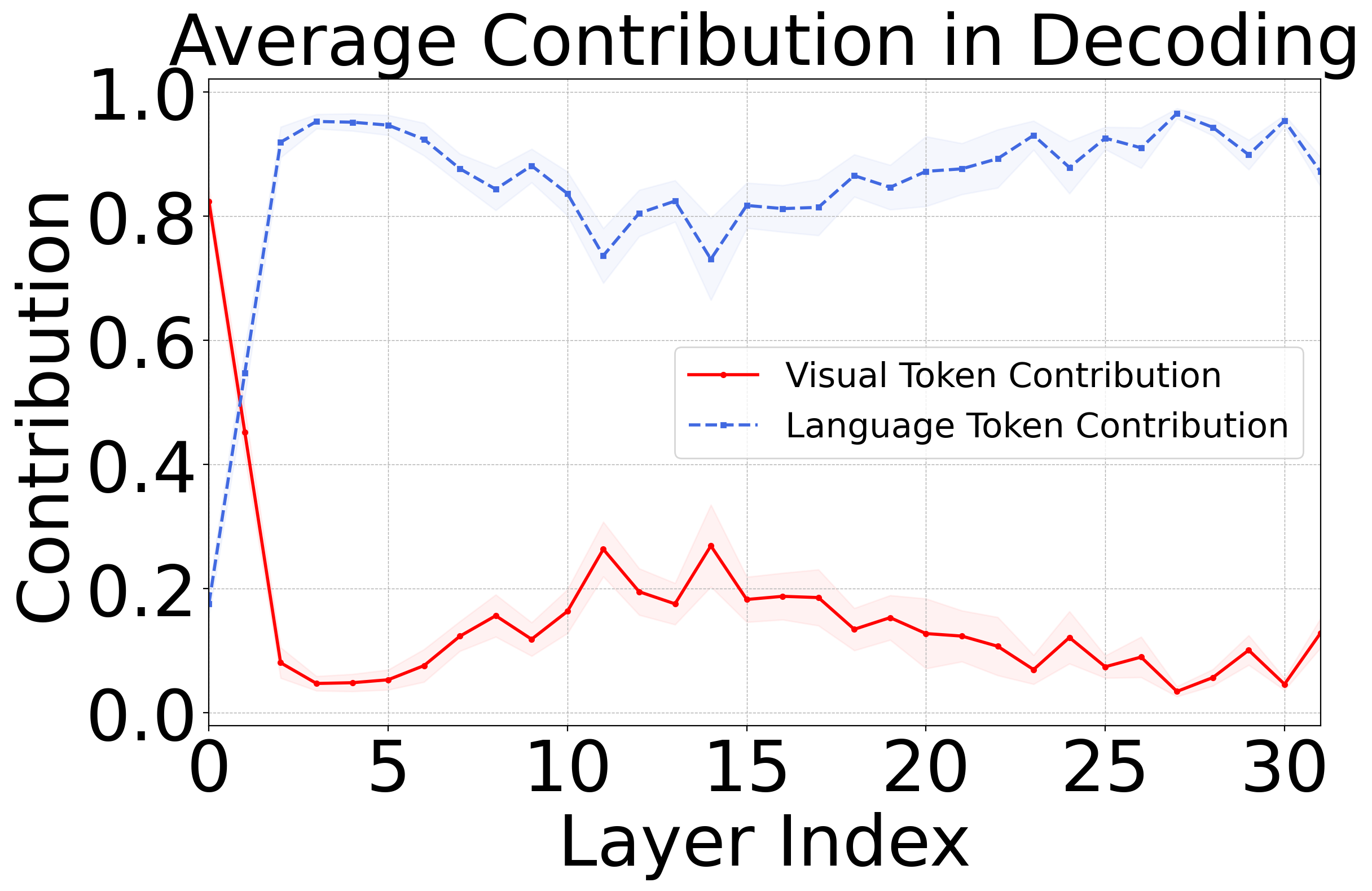}
        \caption{Contribution}
        \label{fig:contribution}
    \end{minipage}%
    \hfill
    \begin{minipage}[t]{0.47\textwidth}
        \centering
        \includegraphics[width=\textwidth]{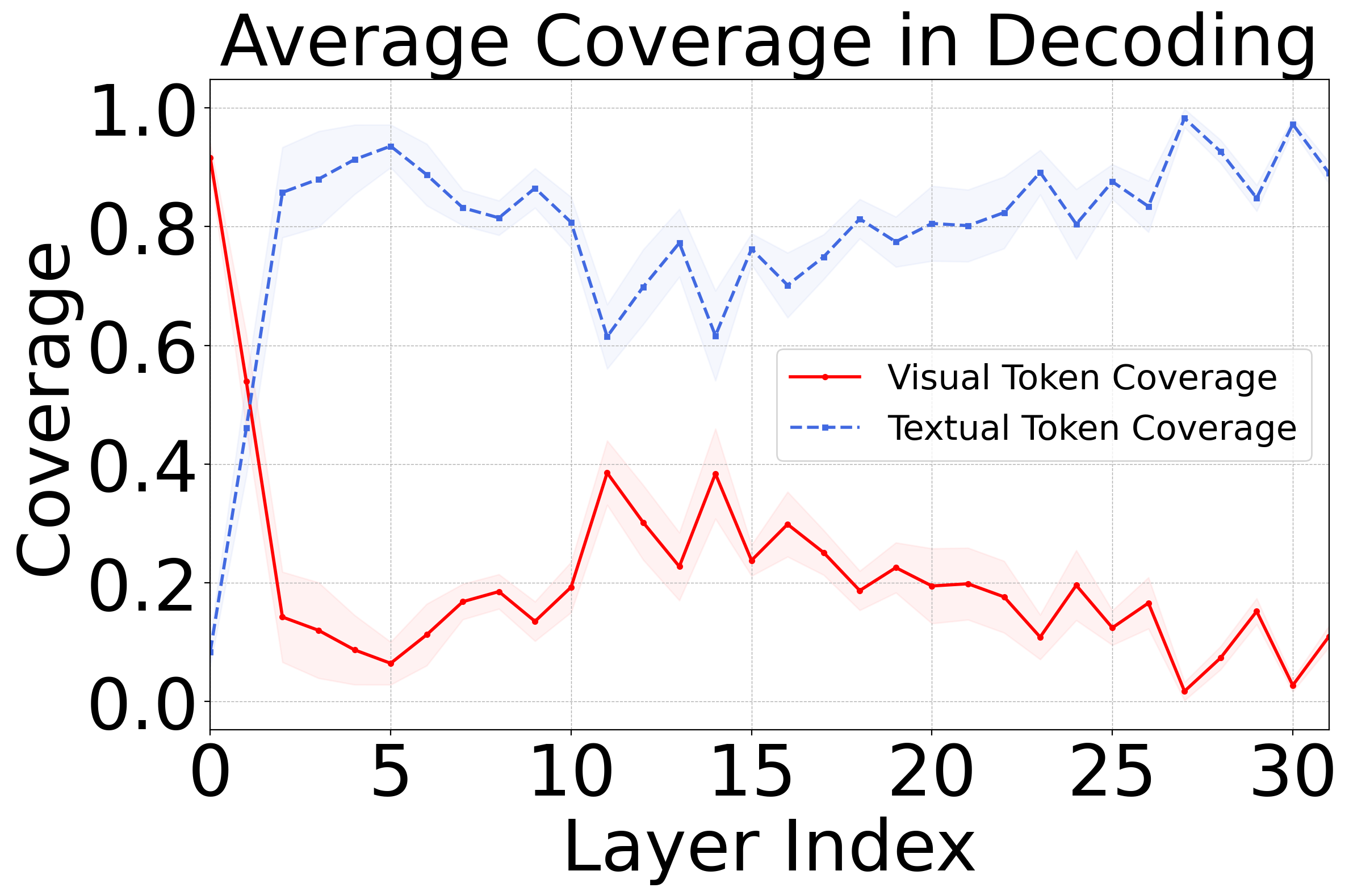}
        \caption{Coverage}
        \label{fig:coverage}
    \end{minipage}
\end{figure}

Here, we use $J_{mod}$ and $J_{all}$ to denote KV cache indices of one selected modality (language or vision) and all modalities respectively in the input prompt. As shown in Figure \ref{fig:contribution}, VLMs allocate primary attention to visual tokens in the first layer, while in the second layer, the contributions of visual and language tokens are comparable. From the third layer onward, language tokens dominate, with a slight increase in visual token contribution in the middle layers. Additionally, we also propose $\Coverage$ to analyze the ratio of the number of tokens from a specific modality, defined as follows:

\begin{equation}
    \Coverage_{mod}^{(l)} \coloneq \frac{1}{T-t+1} \sum_{i \geq t}\frac{\sum_{j \in J_{mod}}{\TopK(A^{(l)}, k)_{ij}}}{\sum_{j \in J_{all}}{\TopK(A^{(l)}, k)_{i j}}}
\end{equation}

Specifically,

\begin{equation}
\TopK(A, k)_{ij} = \mathbbm{1}\lbrack{A_{ij} \in \{ A_{i}^{(r - k + 1)}, A_{i}^{(r - k + 2)},...,A_{i}^{(r)} \} \rbrack}
\end{equation}

Here, we define $k$ as $\floor{\alpha \cdot T}$, $r$ as the number of columns of matrix $A$, and overload notations to use $A_i^{(n)}$ as the $n$-th order statistic of the row vector. Additionally, $\alpha \in (0, 1)$ denotes the token budget threshold (e.g. $\alpha = 10\%$ means only the 10\% of sequence length of KV cache is retained). After KV cache compression with $\TopK$ selection, we assess the ratio of visual attention to language attention among the remaining tokens. In Figure \ref{fig:coverage}, with $\alpha = 10\%$ applied, a similar trend in coverage is observed compared to Figure \ref{fig:contribution}. Notably, in the middle layer, visual tokens constitute a larger proportion than their contribution, indicating that the contribution per token is smaller. 

Furthermore, since contribution of language and visual tokens differ by layer, the optimal budget for each layer may depend on the ratio of visual to language tokens in the prompt. In contrast, previous pyramid-style cache allocation methods do not adapt to the input prompts at inference time.

\end{document}